\documentclass{article}

\usepackage[eandd, final]{neurips_2026}

\usepackage[utf8]{inputenc} 
\usepackage[T1]{fontenc}    
\usepackage[colorlinks,linkcolor=blue,anchorcolor=black,citecolor=blue]{hyperref}
\usepackage{url}            
\usepackage{booktabs}       
\usepackage{amsfonts}       
\usepackage{nicefrac}       
\usepackage{microtype}      
\usepackage{graphicx}
\usepackage[table]{xcolor}
\usepackage{xcolor}         
\definecolor{Gray}{gray}{0.9}
\usepackage{threeparttable}
\usepackage{amssymb}
\usepackage{pifont}

\usepackage{amsmath}
\usepackage{amsthm}
\usepackage{array}
\usepackage{siunitx}
\usepackage{xspace}
\usepackage{longtable}
\usepackage{marvosym}
\usepackage{enumerate}
\usepackage{forest}
\usepackage{mathrsfs}
\usepackage{arydshln}
\usepackage{wrapfig}
\usepackage{tcolorbox}

\newtheorem{definition}{Definition}
\usepackage{makecell}
\usepackage{thmtools} 
\usepackage{thm-restate}

\usepackage{enumitem}
\usepackage{adjustbox}

\makeatletter
\newcommand*\bigcdot{\mathpalette\bigcdot@{.5}}
\newcommand*\bigcdot@[2]{\mathbin{\vcenter{\hbox{\scalebox{#2}{$\m@th#1\bullet$}}}}}
\makeatother

\usepackage[capitalize]{cleveref}
\crefname{section}{Sec.}{Secs.}
\Crefname{section}{Section}{Sections}
\Crefname{table}{Table}{Tables}
\crefname{table}{Tab.}{Tabs.}

\tcbset{
  takeaways/.style={
    colback=blue!3,
    colframe=blue!55!black,
    boxrule=0.6pt,
    arc=2pt,
    left=6pt, right=6pt, top=4pt, bottom=4pt,
    fonttitle=\bfseries,
    title={Takeaways}
  }
}
 
\newcommand{\best}[1]{\textbf{#1}}
\newcommand{\secnd}[1]{\underline{#1}}

\usepackage{multirow}
 
\title{Are We Making Progress in Multimodal Domain Generalization? A Comprehensive Benchmark Study}

\author{
Hao Dong$^{1,2}$ \quad
Hongzhao Li$^{3}$\thanks{Corresponding author.} \quad
Shupan Li$^{3}$  \quad
Muhammad Haris Khan$^{4}$ \\[0.3em]
\textbf{Eleni Chatzi}$^{5}$ \quad
\textbf{Olga Fink}$^{6}$ \\[0.5em]
\normalsize
$^{1}$ELLIS Institute Finland  \quad
$^{2}$Tampere University \quad
$^{3}$Zhengzhou University \quad \\
$^{4}$MBZUAI \quad
$^{5}$ETH Z\"urich \quad
$^{6}$EPFL
}

\begin{document}

\maketitle


\setcounter{footnote}{0}
\begin{abstract}
Despite the growing popularity of Multimodal Domain Generalization (MMDG) for enhancing model robustness, it remains unclear whether reported performance gains reflect genuine algorithmic progress or are artifacts of inconsistent evaluation protocols. Current research is fragmented, with studies varying significantly across datasets, modality configurations, and experimental settings. Furthermore, existing benchmarks focus predominantly on action recognition, often neglecting critical real-world challenges such as input corruptions, missing modalities, and model trustworthiness. This lack of standardization obscures a reliable assessment of the field's advancement. To address this issue, we introduce \textbf{MMDG-Bench}, {the first unified} and comprehensive benchmark for MMDG, which standardizes evaluation across six datasets spanning three diverse tasks: action recognition, mechanical fault diagnosis, and sentiment analysis. MMDG-Bench encompasses six modality combinations, nine representative methods, and multiple evaluation settings. Beyond standard accuracy, it systematically assesses corruption robustness, missing-modality generalization, misclassification detection, and out-of-distribution detection. With $7,402$ neural networks trained in total across $95$ unique cross-domain tasks, MMDG-Bench yields five key findings: (1) under fair comparisons, recent specialized MMDG methods offer only marginal improvements over ERM baseline; (2) no single method consistently outperforms others across datasets or modality combinations; (3) a substantial gap to upper-bound performance persists, indicating that MMDG remains far from solved; (4) trimodal fusion does not consistently outperform the strongest bimodal configurations; and (5) all evaluated methods exhibit significant degradation under corruption and missing-modality scenarios, with some methods further compromising model trustworthiness. We release MMDG-Bench to enable more rigorous, reproducible, and directly comparable evaluation, addressing current limitations in evaluation practices and providing a stronger foundation for future progress in multimodal domain generalization.\footnotemark
\end{abstract}

\footnotetext{\url{https://github.com/lihongzhao99/MMDG_Benchmark}}

\begin{figure}[t]
  \centering  \includegraphics[width=0.98\linewidth]{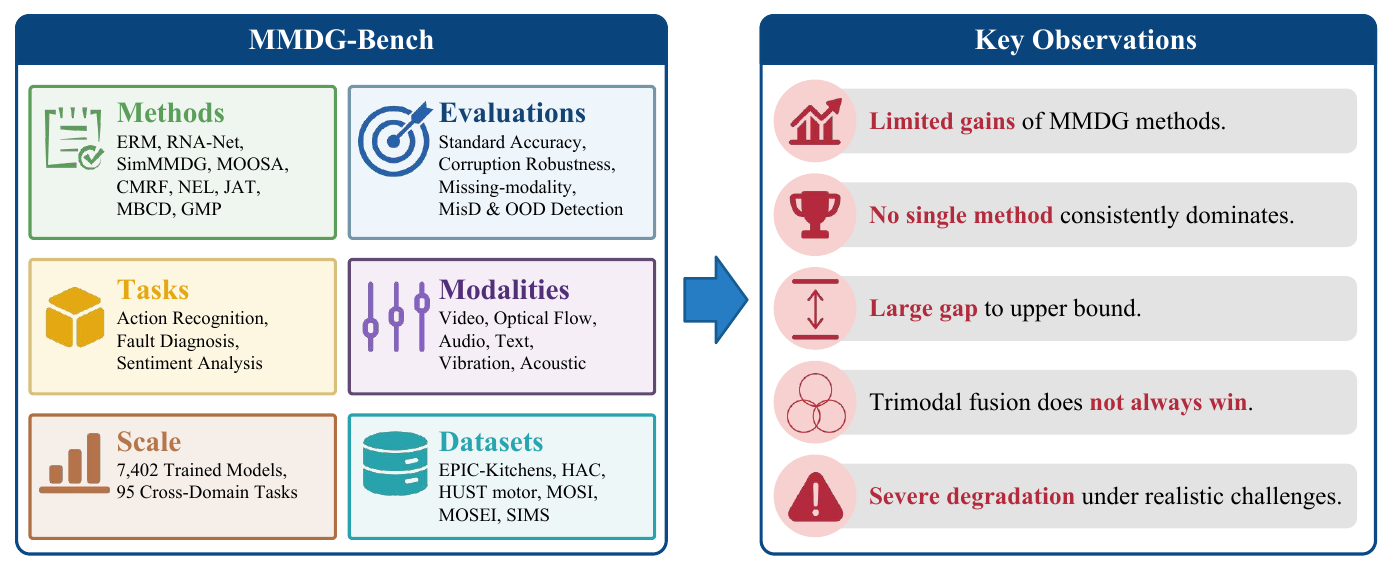}
  \vspace{-0.8em}
   \caption{An overview of the MMDG-Bench and a summary of our key observations.}
   \label{fig:frame}
\end{figure}
  \vspace{-0.8em}

\section{Introduction}
Machine learning (ML) models often suffer substantial performance degradation when deployed in dynamic real-world environments due to distribution shifts between training and testing data~\cite{torralba2011unbiased}. Consequently, generalizing to unseen domains has become a central challenge for building reliable ML systems. Multimodal learning, which integrates complementary signals such as video, audio, and optical flow, is widely regarded as a promising approach to improve robustness. While multimodal models achieve strong in-distribution performance across applications including egocentric action recognition~\cite{Damen2018EPICKITCHENS}, mechanical fault diagnosis~\cite{fink2026physics,fink2026physics2}, and affective computing~\cite{zadeh2016mosi,zadeh2018multimodal,yu2020ch}, they remain brittle under domain shifts caused by environmental changes, operating conditions, or cultural variations. Moreover, multimodal systems introduce unique challenges such as modality imbalance, unreliable fusion, and sensitivity to missing or corrupted inputs~\cite{dong2023SimMMDG,fan2024cross}. These challenges have driven increasing interest in multimodal domain generalization (MMDG), with a growing body of work proposing specialized methods that report consistent empirical gains~\cite{Planamente_2022_WACV,dong2023SimMMDG,dong2024moosa,fan2024cross,zhang2025nonpolarized,li2025towards,wang2025modality,li2026balancing,li2026multimodal}.

Despite this apparent progress, it remains unclear to what extent current MMDG methods yield genuine improvements in cross-domain generalization, as opposed to benefiting from inconsistent evaluation protocols. In unimodal domain generalization, DomainBed~\cite{gulrajani2020search} revealed that \emph{carefully tuned empirical risk minimization (ERM) can match or outperform many specialized methods, fundamentally reshaping the field’s understanding of progress}. In contrast, MMDG lacks a comparable, rigorous benchmark. Existing evaluations vary widely in datasets, modality configurations, training protocols, and metrics, often focusing narrowly on action recognition while overlooking realistic challenges such as missing modalities, input corruptions, and model trustworthiness. Consequently, this lack of standardization hinders reliable assessment and raises a fundamental question: \emph{are we measuring genuine progress, or simply overfitting to biased evaluation protocols?}

To answer this question, we introduce \textbf{MMDG-Bench}, a comprehensive and standardized benchmark for evaluating multimodal domain generalization (Figure~\ref{fig:frame}). MMDG-Bench unifies evaluation across \textbf{six datasets} spanning \textbf{three tasks}: egocentric action recognition (EPIC-Kitchens~\cite{Damen2018EPICKITCHENS}, HAC~\cite{dong2023SimMMDG}), mechanical fault diagnosis (HUST Motor~\cite{zhao2024domain}), and multimodal sentiment analysis (CMU-MOSI~\cite{zadeh2016mosi}, CMU-MOSEI~\cite{zadeh2018multimodal}, CH-SIMS~\cite{yu2020ch}). It covers \textbf{six modality combinations} and evaluates \textbf{nine representative methods} across $95$ cross-domain tasks under both multi-source and single-source settings. Beyond standard accuracy, we systematically assess corruption robustness, missing-modality generalization, misclassification detection, and out-of-distribution (OOD) detection, capturing both predictive performance and model reliability. To ensure fair comparison, we standardize data splits, hyperparameter search, optimization protocols, and model selection criteria. With $7,402$ neural networks trained in total, MMDG-Bench provides a comprehensive evaluation and yields critical insights to guide future research:
\begin{itemize}[leftmargin=*]
\setlength\itemsep{0em}
\item Under fair evaluation, specialized MMDG methods offer only marginal gains over strong baselines, with ERM frequently matching or outperforming recent approaches.
\item No single method consistently dominates across datasets or modality configurations.
\item A substantial gap relative to the Oracle model remains, confirming that MMDG is far from solved.
\item Trimodal fusion does not consistently surpass the strongest bimodal configurations, challenging the assumption that additional modalities inherently improve generalization.
\item All methods remain highly vulnerable to corruptions and missing modalities, with some degrading model trustworthiness despite improving raw accuracy.
\end{itemize}
These results suggest that progress in MMDG may be partially overestimated due to inconsistencies in evaluation protocols, underscoring the need for rigorous and standardized benchmarking.

\section{A Comprehensive Benchmark for Multimodal Domain Generalization}
\label{sec:bench}

This section outlines the design and scope of MMDG-Bench. We first formalize the relevant MMDG paradigms (Sec.~\ref{sec:paradigms}), then describe the representative methods included (Sec.~\ref{sec:methods}), and finally detail the datasets, modality configurations, backbone architectures, evaluation protocols, and hyperparameter search procedures utilized (Sec.~\ref{sec:setup}).

\subsection{Multimodal Domain Generalization Paradigms}
\label{sec:paradigms}

Let $\mathcal{M} = \{m_1, \dots, m_K\}$ denote a set of $K$ modalities (e.g., video, audio, optical flow). A multimodal sample $(x^{m_1}, \dots, x^{m_K}, y)$ is drawn from a joint distribution $P_{\mathcal{D}}$ associated with domain $\mathcal{D}$, where $x^{m_k}$ represents the input from modality $m_k$, and $y \in \mathcal{Y}$ is the corresponding label.

\begin{definition}[Multi-source MMDG]
Given $N_s$ labeled source domains $\{\mathcal{D}^s_i\}_{i=1}^{N_s}$ sharing a common label space and modality set, multi-source MMDG seeks to learn a model 
$f: \mathcal{X}^{m_1} \times \cdots \times \mathcal{X}^{m_K} \rightarrow \mathcal{Y}$ 
that generalizes effectively to an unseen target domain $\mathcal{D}^t$, without access to any target-domain data during training.
\end{definition}

\begin{definition}[Single-source MMDG]
Given a single labeled source domain $\mathcal{D}^s$ and an unseen target domain $\mathcal{D}^t$ sharing the same label space and modality set, single-source MMDG seeks to train a model that transfers robustly from $\mathcal{D}^s$ to $\mathcal{D}^t$ without target-domain access during training.
\end{definition}

\begin{definition}[Corruption Robustness]
Given a source-trained MMDG model, corruption robustness evaluates performance when one or more target-domain modalities undergo realistic perturbations (e.g., audio wind noise, video defocus blur). It is quantified by the performance degradation between clean and corrupted target conditions.
\end{definition}

\begin{definition}[Missing-modality Generalization]
Given a source-trained MMDG model, this setting measures generalization when modalities present during training are absent during target-domain inference, reflecting real-world scenarios such as sensor failures or incomplete observations.
\end{definition}
 
\subsection{Multimodal Domain Generalization Methods}
\label{sec:methods}

MMDG-Bench evaluates nine representative MMDG methods alongside an Oracle reference. 

\textbf{ERM}~\cite{vapnik1999overview} serves as our foundational baseline, pooling all source domains to minimize empirical risk without explicit MMDG objectives.

\textbf{RNA-Net}~\cite{Planamente_2022_WACV} aligns the average feature norms across modalities using a Relative Norm Alignment objective, mitigating modality-induced domain bias without requiring domain annotations.

\textbf{SimMMDG}~\cite{dong2023SimMMDG} decomposes representations into modality-shared and modality-specific components. It uses supervised contrastive learning to extract domain-invariant shared features and incorporates a cross-modal translation module to improve missing-modality robustness.

\textbf{MOOSA}~\cite{dong2024moosa} utilizes masked cross-modal translation and multimodal jigsaw puzzles as self-supervised auxiliary tasks, combined with entropy-guided modality balancing. Though designed for open-set MMDG, it remains highly competitive in standard closed-set settings.

\textbf{CMRF}~\cite{fan2024cross} addresses modality competition and inconsistent unimodal flatness in sharpness-aware minimization. It flattens the cross-modal representation landscape by interpolating between modality-specific minima, followed by feature distillation into individual modality branches.

\textbf{NEL}~\cite{zhang2025nonpolarized} mitigates representation polarization, where one modality dominates the shared embedding space, via a nonpolarized learning objective that encourages balanced, domain-invariant multimodal representations.

\textbf{JAT}~\cite{li2025towards} performs adversarial training using gradient reversal layers on both modality-specific and fused representations, enforcing domain invariance across multiple representation levels.

\textbf{MBCD}~\cite{wang2025modality} observes that asynchronous modality convergence limits conventional weight averaging and introduces a collaborative distillation framework utilizing adaptive modality dropout, gradient consistency regularization, and an EMA teacher for cross-modal knowledge transfer.

\textbf{GMP}~\cite{li2026balancing} revisits gradient modulation under domain shift by decomposing modality gradients into classification-oriented and domain-invariant components. By dynamically modulating and projecting these gradients based on semantic and domain confidence, it resolves optimization conflicts.

Finally, our \textbf{Oracle} model is trained directly on target-domain data. While not a valid domain generalization method, it provides an empirical performance ceiling to quantify the remaining gap between current MMDG approaches and ideal target-domain performance.
 
\subsection{Experimental Setups}
\label{sec:setup}

\begin{figure}[t]
  \centering  \includegraphics[width=0.8\linewidth]{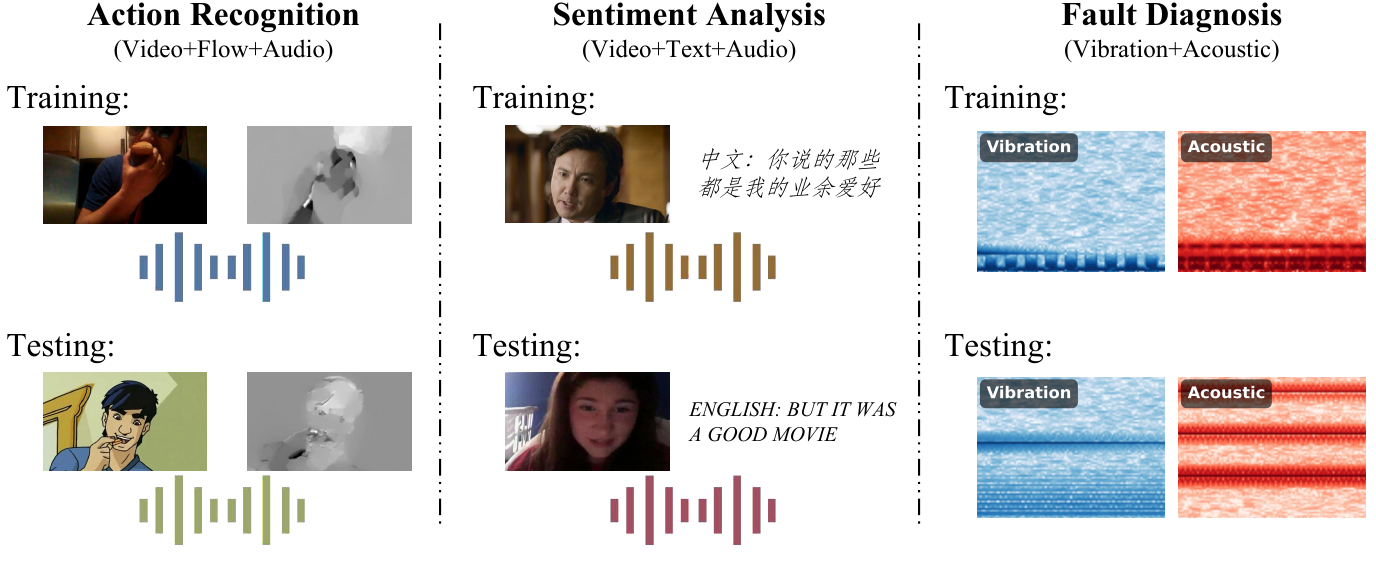}
  \vspace{-1em}
   \caption{Illustration of three core tasks included in the MMDG-Bench. 
   }
   \label{fig:mmdg_example}
\end{figure}
 
\textbf{Datasets.} MMDG-Bench unifies six datasets across three task families for diverse evaluation (Figure~\ref{fig:mmdg_example}). 
For \textit{\textbf{action recognition}}, we include {EPIC-Kitchens}~\cite{Damen2018EPICKITCHENS} (eight classes across three kitchen environments) and {HAC}~\cite{dong2023SimMMDG} (seven classes performed by humans, animals, and cartoons). Both provide video (V), audio (A), and optical flow (F).
For \textit{\textbf{mechanical fault diagnosis}}, we adopt {HUST motor}~\cite{zhao2024domain}, comprising four operating-condition domains with vibration and acoustic signals.
For \textit{\textbf{sentiment analysis}}, we evaluate {CMU-MOSI}~\cite{zadeh2016mosi}, {CMU-MOSEI}~\cite{zadeh2018multimodal}, and {CH-SIMS}~\cite{yu2020ch} (video, audio, text); each acts as a distinct domain for cross-dataset MMDG. Detailed statistics, preprocessing, and splits are in the Appendix~\ref{sec:dataset}.

\textbf{Modality combinations.} We assess six modality configurations: four for action recognition (V+A, V+F, A+F, V+A+F), one for fault diagnosis (vibration+acoustic), and one for sentiment analysis (video+audio+text), enabling systematic evaluation of both bimodal and trimodal fusion.

\textbf{Backbone architectures.} For action recognition, we build on MMAction2~\cite{2020mmaction2}: video via Kinetics-400 pretrained SlowFast~\cite{Feichtenhofer_2019_ICCV}, audio via VGGSound pretrained ResNet-18~\cite{7780459}, and optical flow via a Kinetics-initialized SlowFast slow-only pathway. For fault diagnosis, we employ a four-layer 1D CNN for vibration and acoustic signals~\cite{zhao2024domain}. For sentiment analysis~\cite{guo2025bridging}, we extract 768-dimensional text embeddings via pretrained BERT~\cite{devlin2019bert}, audio features via LibROSA~\cite{mcfee2015librosa}, and visual facial features via OpenFace 2.0~\cite{baltruvsaitis2016openface}, fused by a Transformer encoder~\cite{46201}. 

\textbf{Evaluation protocols.} Multi-source MMDG follows a leave-one-domain-out protocol, while single-source evaluates all source-target pairs. For sentiment analysis, we report binary accuracy (ACC2), F1 score, and mean absolute error (MAE). To ensure fair comparisons, all methods use identical data splits, optimizers, and training-domain validation for model selection~\citep{gulrajani2020search}.

\textbf{Hyperparameter search.} For each algorithm-dataset pair, we evaluate the default hyperparameters alongside 10 random-search trials (detailed in the Appendix~\ref{sec:hyper}). The optimal configuration, selected via training-domain validation, is retrained with two additional random seeds to mitigate variance from random initialization and stochastic optimization, and the final performance is reported as the average across all seeds to provide a more reliable estimate. 
This rigorous protocol requires training $7,402$ neural networks, making MMDG-Bench the most comprehensive MMDG benchmark study to date.

\begin{table*}[t!]
\centering
\caption{Multimodal {multi-source} DG with different modality combinations on EPIC-Kitchens and HAC datasets for \textbf{action recognition} task.}
\vspace{0.4em}

\setlength{\tabcolsep}{2.5pt}
\resizebox{\linewidth}{!}{
\begin{threeparttable}
\begin{tabular}{lcccccccccccc}
\toprule
& \multicolumn{3}{c}{\textbf{Modality}} & \multicolumn{4}{c}{\textbf{EPIC-Kitchens dataset}}& \multicolumn{4}{c}{\textbf{HAC dataset}}\\
\cmidrule(lr){2-4} \cmidrule(lr){5-8} \cmidrule(lr){9-12} 
\textbf{Method} & Video & Audio & Flow & D2, D3 $\rightarrow$ D1 & D1, D3 $\rightarrow$ D2 & D1, D2 $\rightarrow$ D3 & \textit{Mean} & A, C $\rightarrow$ H & H, C $\rightarrow$ A & H, A $\rightarrow$ C & \textit{Mean}\\

\midrule

ERM & $\checkmark$& $\checkmark$&    & 57.47 & 61.20 & 60.68 & 59.78 & 75.91 & 77.48 & 53.40 & 68.93 \\
RNA-Net~\cite{Planamente_2022_WACV} & $\checkmark$& $\checkmark$&  & 57.24 & 60.40 & 60.47 & 59.37 & 75.20 & 77.48 & 53.58 & 68.75 \\
SimMMDG~\cite{dong2023SimMMDG}& $\checkmark$& $\checkmark$&   & 58.62 & 66.40 & 65.30 & 63.44 & \secnd{78.59} & 78.04 & \best{55.79} & \secnd{70.81} \\
MOOSA~\cite{dong2024moosa}& $\checkmark$& $\checkmark$&   & \secnd{59.31} & 65.33 & \best{66.63} & \secnd{63.76} & \best{79.38} & \secnd{78.70} & \secnd{54.78} & \best{70.95} \\
CMRF~\cite{fan2024cross}& $\checkmark$& $\checkmark$&   & 57.01 & \secnd{69.47} & 64.37 & 63.62 & 77.94 & 78.26 & 51.84 & 69.35 \\
NEL~\cite{zhang2025nonpolarized}& $\checkmark$& $\checkmark$&  & 54.63 & 66.75 & 62.55 & 61.31 & 76.33 & 76.42 & 51.07 & 67.94 \\
JAT~\cite{li2025towards}& $\checkmark$& $\checkmark$&   & 57.98 & 66.82 & 64.14 & 62.98 & 78.16 & 77.99 & 53.11 & 69.75 \\
MBCD~\cite{wang2025modality}& $\checkmark$& $\checkmark$&   & \best{59.38} & \best{69.60} & \secnd{65.63} & \best{64.87} & 78.12 & \best{78.91} & 53.49 & 70.17 \\
GMP~\cite{li2026balancing} & $\checkmark$& $\checkmark$&  & 57.62 & 65.39 & 64.88 & 62.63 & 77.36 & 76.47 & 52.33 & 68.72 \\
\textit{Oracle} & $\checkmark$& $\checkmark$&  & \textit{60.23} & \textit{76.13} & \textit{76.80} & \textit{71.05} & \textit{92.75} & \textit{97.16} & \textit{88.53} & \textit{92.81} \\

\midrule

ERM & $\checkmark$& &$\checkmark$ & 59.77 & 66.13 & 62.73 & 62.88 & 76.93 & \best{77.59} & 49.82 & 68.11 \\
RNA-Net~\cite{Planamente_2022_WACV} & $\checkmark$& & $\checkmark$& 60.00 & 67.47 & 64.58 & 64.02 & 77.58 & 76.71 & 52.85 & 69.05 \\
SimMMDG~\cite{dong2023SimMMDG} & $\checkmark$& & $\checkmark$& 60.69 & \secnd{69.33} & 64.07 & 64.70 & 78.95 & 75.94 & \secnd{54.60} & 69.83 \\
MOOSA~\cite{dong2024moosa}& $\checkmark$& &  $\checkmark$ & 61.84 & 69.20 & 64.89 & 65.31 & 80.46 & 76.71 & \best{56.71} & \best{71.29} \\
CMRF~\cite{fan2024cross}& $\checkmark$& & $\checkmark$  & 61.61 & \secnd{69.33} & 65.81 & \secnd{65.58} & \best{81.47} & 76.38 & 52.30 & 70.05 \\
NEL~\cite{zhang2025nonpolarized}& $\checkmark$& &   $\checkmark$& 59.00 & 67.02 & 63.99 & 63.34 & 80.29 & 76.45 & 51.16 & 69.30 \\
JAT~\cite{li2025towards}& $\checkmark$& &   $\checkmark$& \secnd{61.88} & 68.79 & \secnd{65.82} & 65.50 & 78.39 & \secnd{77.38} & 52.17 & 69.31 \\
MBCD~\cite{wang2025modality}& $\checkmark$& & $\checkmark$  & \best{63.36} & \best{71.06} & \best{67.18} & \best{67.20} & \secnd{81.39} & 77.08 & 53.67 & \secnd{70.71} \\
GMP~\cite{li2026balancing} & $\checkmark$& & $\checkmark$ & 60.37 & 67.21 & \secnd{65.82} & 64.47 & 77.92 & 76.35 & 52.56 & 68.94 \\
\textit{Oracle} & $\checkmark$& & $\checkmark$ & \textit{65.52} & \textit{80.00} & \textit{81.21} & \textit{75.58} & \textit{93.48} & \textit{96.59} & \textit{85.78} & \textit{91.95} \\

\midrule

ERM & & $\checkmark$&$\checkmark$ & 52.18 & 61.47 & 58.31 & 57.32 & 55.66 & 63.90 & \secnd{47.24} & 55.60 \\
RNA-Net~\cite{Planamente_2022_WACV} & & $\checkmark$&$\checkmark$ & 52.41 & 59.47 & 62.53 & 58.14 & 56.67 & 64.13 & 46.42 & 55.74 \\
SimMMDG~\cite{dong2023SimMMDG} & & $\checkmark$&$\checkmark$ & 55.86 & \best{69.20} & \secnd{63.04} & 62.70 & 58.83 & 65.45 & 45.96 & 56.75 \\
MOOSA~\cite{dong2024moosa}& & $\checkmark$&  $\checkmark$ & \best{58.16} & 68.27 & 62.42 & \best{62.95} & \secnd{59.55} & \best{66.11} & 46.88 & \secnd{57.51} \\
CMRF~\cite{fan2024cross}&  & $\checkmark$&   $\checkmark$ & 53.56 & \secnd{68.40} & 61.81 & 61.26 & 58.54 & 65.34 & 46.42 & 56.77 \\
NEL~\cite{zhang2025nonpolarized}&  & $\checkmark$& $\checkmark$  & 56.24 & 63.33 & 61.09 & 60.22 & 58.80 & 64.08 & 45.95 & 56.28 \\
JAT~\cite{li2025towards}&  & $\checkmark$& $\checkmark$  & \secnd{56.83} & 65.26 & 62.17 & 61.42 & 59.32 & 65.12 & 45.07 & 56.50 \\
MBCD~\cite{wang2025modality}& & $\checkmark$&   $\checkmark$ & 56.78 & 66.57 & \best{65.36} & \secnd{62.90} & \best{61.60} & \secnd{66.07} & \best{48.71} & \best{58.79} \\
GMP~\cite{li2026balancing} & & $\checkmark$& $\checkmark$ & 55.38 & 64.92 & 62.77 & 61.02 & 57.31 & 65.13 & 46.84 & 56.43 \\
\textit{Oracle} & & $\checkmark$& $\checkmark$ & \textit{59.77} & \textit{74.13} & \textit{73.61} & \textit{69.17} & \textit{81.52} & \textit{90.91} & \textit{68.35} & \textit{80.26} \\

\midrule

ERM & $\checkmark$& $\checkmark$& $\checkmark$ & 56.78 & 66.67 & 65.61 & 63.02 & 73.32 & 76.49 & 53.86 & 67.89 \\
RNA-Net~\cite{Planamente_2022_WACV} & $\checkmark$& $\checkmark$& $\checkmark$ & 57.24 & 66.00 & 67.97 & 63.74 & 73.68 & 76.16 & 54.41 & 68.08 \\
SimMMDG~\cite{dong2023SimMMDG} & $\checkmark$& $\checkmark$& $\checkmark$ & \best{63.91} & 71.47 & 68.89 & \best{68.09} & 78.15 & 75.39 & 54.60 & 69.38 \\
MOOSA~\cite{dong2024moosa}& $\checkmark$& $\checkmark$& $\checkmark$  & 59.77 & \best{72.93} & \best{69.82} & \secnd{67.51} & 75.70 & 78.37 & \best{56.43} & 70.17 \\
CMRF~\cite{fan2024cross}& $\checkmark$& $\checkmark$&$\checkmark$   & \secnd{62.76} & 70.40 & 68.17 & 67.11 & \secnd{79.02} & \best{80.35} & 54.87 & \best{71.41} \\
NEL~\cite{zhang2025nonpolarized}& $\checkmark$& $\checkmark$&  $\checkmark$ & 60.46 & 68.48 & 65.02 & 64.65 & 77.26 & 78.10 & \secnd{55.88} & 70.41 \\
JAT~\cite{li2025towards}& $\checkmark$& $\checkmark$& $\checkmark$  & 61.38 & 69.96 & 66.37 & 65.90 & 77.32 & 77.59 & 54.88 & 69.93 \\
MBCD~\cite{wang2025modality}& $\checkmark$& $\checkmark$& $\checkmark$  & 61.29 & 71.24 & \secnd{69.50} & 67.34 & \best{79.06} & \secnd{79.21} & 55.64 & \secnd{71.30} \\
GMP~\cite{li2026balancing} & $\checkmark$& $\checkmark$& $\checkmark$ & 59.77 & 68.39 & 66.33 & 64.83 & 78.26 & 77.35 & 53.97 & 69.86 \\
\textit{Oracle} & $\checkmark$& $\checkmark$& $\checkmark$ & \textit{65.52} & \textit{79.47} & \textit{78.64} & \textit{74.54} & \textit{92.75} & \textit{96.02} & \textit{86.24} & \textit{91.67} \\

\bottomrule
\end{tabular}

\end{threeparttable}
}

\label{tab:epic-hac-dg}
\end{table*}


\section{Multimodal Domain Generalization Under Fair Comparison}

\label{sec:fair}
\textbf{Experimental setup.} This section examines whether recent MMDG algorithms still outperform strong baselines once major confounding factors are removed. To ensure a fair and rigorous comparison, we standardize all key pipeline components, including data splits, batch sizes, optimizers, and model selection strategies. All methods are selected using training-domain validation, thereby isolating algorithmic contributions rather than evaluation artifacts.

\textbf{Results on action recognition.} Table~\ref{tab:epic-hac-dg} summarizes multi-source MMDG results on EPIC-Kitchens and HAC. Crucially, \textit{no single method consistently dominates across datasets, modality combinations, or domain shifts}. Performance rankings fluctuate substantially, and gains over strong baselines (e.g., ERM, SimMMDG) are often modest, indicating that reported MMDG progress remains highly context-dependent. Furthermore, the Audio+Flow configuration consistently yields the weakest results across both benchmarks, confirming that video remains the most informative modality for action recognition.

\textbf{Results on fault diagnosis.} Table~\ref{tab:hust-dg} presents results multi-source MMDG on HUST motor. The performance gap across methods is larger than that observed in action recognition. MOOSA achieves the highest mean accuracy ($78.23\%$), followed by GMP and CMRF, significantly outperforming ERM ($69.90\%$). 
However, the ranking of methods differs from that in action recognition. MBCD performs strongly on EPIC-Kitchens but drops to the lowest rank on HUST, while GMP improves from a mid-tier position in action recognition to second place here. These drastic ranking shifts reveal that current methods fail to generalize reliably across task families, highlighting the risk of drawing broad conclusions from limited benchmark settings.

\begin{table*}[t!]
\centering
\caption{Multimodal multi-source DG on HUST motor dataset with vibration and acoustic modalities for \textbf{fault diagnosis} task.}
\vspace{0.4em}
\resizebox{0.9\linewidth}{!}{
\begin{threeparttable}
\begin{tabular}{lcccccccccccc}
\toprule
\textbf{Method} & D2, D3, D4 $\rightarrow$ D1 & D1, D3, D4 $\rightarrow$ D2 &  D1, D2, D4 $\rightarrow$ D3 &  D1, D2, D3 $\rightarrow$ D4   & \textit{Mean} \\

\midrule

ERM      & 42.25 & 83.92 & 76.25 & 77.17 & 69.90 \\
RNA-Net~\cite{Planamente_2022_WACV}  & 43.50 & 84.58 & 73.25 & 79.58 & 70.23 \\
SimMMDG~\cite{dong2023SimMMDG}  & 42.33 & 88.50 & 82.42 & 82.08 & 73.83 \\
MOOSA~\cite{dong2024moosa}    & \best{51.08} & \best{93.00} & \secnd{84.92} & \best{83.92} & \best{78.23} \\
CMRF~\cite{fan2024cross}     & \secnd{47.42} & 87.92 & 83.67 & 80.75 & 74.94 \\
NEL~\cite{zhang2025nonpolarized}      & 46.97 & 80.50 & 76.53 & 78.19 & 70.55 \\
JAT~\cite{li2025towards}      & 44.22 & 82.36 & 77.36 & 79.58 & 70.88 \\
MBCD~\cite{wang2025modality}     & 42.89 & 83.72 & 79.31 & 70.64 & 69.14 \\
GMP~\cite{li2026balancing}      & 47.45 & \secnd{91.66} & \best{89.17} & \secnd{81.61} & \secnd{77.47} \\
\textit{Oracle} & \textit{99.83} & \textit{99.83} & \textit{100.00} & \textit{99.83} & \textit{99.87} \\


\bottomrule
\end{tabular}

\end{threeparttable}
}
\vspace{-1em}
\label{tab:hust-dg}
\end{table*}

\textbf{Results on sentiment analysis.} Table~\ref{tab:msa-dg} reports performance multi-source MMDG on sentiment analysis datasets, further highlighting the limitations of current methods. The strongest specialized method (MOOSA, $66.60\%$ ACC2) outperforms ERM ($65.63\%$) by less than one percentage point. In half of the scenarios, ERM matches or exceeds specialized approaches. Moreover, several prominent methods (SimMMDG, MBCD, GMP) underperform ERM on mean ACC2, indicating potential negative transfer in text-centric tasks. Moreover, most methods \textit{perform poorly on regression tasks}, as reflected by high MAE. Ultimately, these results show that current MMDG techniques are highly task-dependent and lack broad cross-domain robustness.

\begin{table*}[t!]
\centering
\caption{Multimodal multi-source DG on MOSI, MOSEI, and SIMS datasets with video, audio, and text modalities for \textbf{sentiment analysis} task.}
\vspace{0.4em}
\resizebox{0.85\linewidth}{!}{
\begin{threeparttable}
\begin{tabular}{l ccc ccc ccc} 
\toprule
\textbf{Method} & \multicolumn{3}{c}{MOSI, MOSEI $\rightarrow$ SIMS} & \multicolumn{3}{c}{MOSI, SIMS $\rightarrow$ MOSEI} & \multicolumn{3}{c}{\textit{Mean}} \\
\cmidrule(lr){2-4} \cmidrule(lr){5-7} \cmidrule(lr){8-10}
& MAE$\downarrow$ & F1$\uparrow$ & ACC2$\uparrow$ & MAE$\downarrow$ & F1$\uparrow$ & ACC2$\uparrow$ & MAE$\downarrow$ & F1$\uparrow$ & ACC2$\uparrow$ \\

\midrule
ERM & \best{1.82} & 69.00 & 63.90 & 1.02 & 67.35 & 67.35 & 1.42 & 68.18 & 65.63 \\
RNA-Net~\cite{Planamente_2022_WACV} & \secnd{1.83} & 66.71 & 64.55 & \secnd{0.92} & 67.22 & 67.22 & \secnd{1.38} & 66.97 & 65.89 \\
SimMMDG~\cite{dong2023SimMMDG} & 1.84 & 64.39 & 61.71 & 1.00 & \secnd{67.87} & 67.65 & 1.42 & 66.13 & 64.68 \\
MOOSA~\cite{dong2024moosa} & 1.89 & \secnd{71.76} & \best{66.30} & 0.96 & 67.17 & 66.90 & 1.43 & \secnd{69.47} & \best{66.60} \\
CMRF~\cite{fan2024cross} & \secnd{1.83} & \best{72.12} & \secnd{65.21} & \best{0.89} & 67.75 & \secnd{67.74} & \best{1.36} & \best{69.94} & \secnd{66.48} \\
NEL~\cite{zhang2025nonpolarized} & 1.91 & 51.79 & 52.44 & 0.99 & 67.55 & 67.52 & 1.45 & 59.67 & 59.98 \\
JAT~\cite{li2025towards} & 1.85 & 67.16 & 64.40 & 0.98 & \best{67.90} & \best{67.87} & 1.42 & 67.53 & 66.14 \\
MBCD~\cite{wang2025modality} & 1.84 & 58.12 & 57.84 & 1.03 & 67.47 & 66.83 & 1.44 & 62.80 & 62.34 \\
GMP~\cite{li2026balancing} & 1.93 & 58.68 & 57.54 & 1.09 & 67.32 & 67.16 & 1.51 & 62.00 & 62.35 \\
\textit{Oracle} & \textit{1.32} & \textit{76.80} & \textit{76.80} & \textit{0.58} & \textit{73.89} & \textit{73.63} & \textit{0.95} & \textit{75.35} & \textit{75.22} \\


\bottomrule
\end{tabular}
\end{threeparttable}
}
\label{tab:msa-dg}
\end{table*}

\textbf{Single-source DG.} Single-source DG results largely reinforce the trends observed in the multi-source setting. On EPIC-Kitchens (Table~\ref{tab:epic-ssdg-va} and Table~\ref{tab:epic-ssdg}), MBCD achieves the best average performance across modality combinations, with SimMMDG and MOOSA closely following. On HAC (Table~\ref{tab:hac-ssdg-appen}), SimMMDG leads in the trimodal V+A+F setting ($63.60\%$), while MBCD remains highly competitive ($63.53\%$). HUST Motor (Table~\ref{tab:hust-ssdg}) provides a particularly challenging evaluation, where limiting training to a single source domain substantially reduces performance for all methods. In severe transfer scenarios (e.g., D1 $\rightarrow$ D4), accuracy declines sharply to $1.75\%-18.14\%$, indicating that existing methods depend heavily on source-domain diversity. This suggests that much of the improvement in multi-source DG may arise from broader source coverage rather than fundamental algorithmic advances. For sentiment analysis (Table~\ref{tab:msa-ssdg}), SimMMDG achieves the strongest average classification performance (F1 and ACC2), while CMRF performs best on MAE.

\textbf{Trimodal fusion does not consistently improve generalization.} Multimodal learning is often assumed to improve robustness by incorporating additional modalities. However, the trimodal (V+A+F) results in Table~\ref{tab:epic-hac-dg} present a more complex picture. On HAC, V+A+F outperforms V+F in only five of nine methods. For several approaches, including ERM, RNA-Net, SimMMDG, and MOOSA, adding a third modality yields minimal benefit or even degrades performance (e.g., MOOSA declines from $71.29\%$ to $70.17\%$).
Methods explicitly designed to address modality competition, such as CMRF, MBCD, and GMP, demonstrate more consistent gains from trimodal integration ($+1.36\%$, $+0.59\%$, $+0.92\%$, respectively), supporting the view that modality competition is a key optimization bottleneck. Nevertheless, current solutions remain only partially effective and fail to deliver substantial, reliable improvements across datasets.

\textbf{Massive gap to Oracle model.} Across all datasets, Oracle results reveal a substantial gap between current MMDG performance and achievable target-domain accuracy. For example, on HAC (V+A), the Oracle reaches $92.81\%$ mean accuracy, surpassing the best-performing method (MOOSA, $70.95\%$) by nearly $22$ percentage points. These results demonstrate that MMDG remains an open and challenging problem and highlight the need for fundamentally new approaches to close this large generalization gap. Appendix~\ref{sec:add-ar} corroborates these conclusions with transformer backbones, foundation-model baselines, per-class accuracy, and an effective-robustness analysis.

\begin{tcolorbox}[takeaways]
\begin{enumerate}[leftmargin=*]
\setlength\itemsep{0em}
\item Under standardized and fair evaluation, specialized MMDG methods yield only marginal improvements over strong baselines.
\item No single method consistently dominates across different datasets, modality combinations, or task families.
\item Trimodal fusion delivers inconsistent gains; while methods targeting modality competition benefit more reliably, current solutions remain incomplete.
\item A substantial performance gap relative to the Oracle persists across all datasets, confirming that MMDG remains far from a solved problem.
\end{enumerate}
\end{tcolorbox}

\begin{table*}[t!]
\centering
\caption{Multimodal \textbf{single-source} DG  with video and audio on EPIC-Kitchens dataset.}
\vspace{0.4em}
\resizebox{0.8\linewidth}{!}{
\begin{threeparttable}
\begin{tabular}{lccccccccccc}
\toprule
& \multicolumn{2}{c}{\textbf{Source: D1}}& \multicolumn{2}{c}{\textbf{Source: D2}}& \multicolumn{2}{c}{\textbf{Source: D3}}\\
\cmidrule(lr){2-3} \cmidrule(lr){4-5} \cmidrule(lr){6-7} 
\textbf{Method} & D1$\rightarrow$ D2 & D1$\rightarrow$ D3 & D2 $\rightarrow$ D1 & D2 $\rightarrow$ D3& D3$\rightarrow$ D1& D3$\rightarrow$ D2  & \textit{Mean}\\
\midrule

ERM     & 51.07 & \secnd{54.72} & 43.45 & 55.44 & 46.67 & 56.13 & 51.25 \\
RNA-Net~\cite{Planamente_2022_WACV}   & 52.53 & 51.85 & 51.03 & 56.26 & 53.79 & 55.60 & 53.51 \\
SimMMDG~\cite{dong2023SimMMDG}  & 53.33 & 51.54 & \secnd{51.72} & 60.16 & \secnd{55.63} & \secnd{58.93} & \secnd{55.22} \\
MOOSA~\cite{dong2024moosa}  & 53.60 & 51.23 & 47.82 & \secnd{61.91} & \best{56.55} & 58.80 & 54.98 \\
CMRF~\cite{fan2024cross}  & \best{58.67} & 51.33 & 49.66 & \best{62.01} & 50.11 & 57.73 & 54.92 \\
NEL~\cite{zhang2025nonpolarized}   & 54.66 & 54.07 & 47.81 & 59.13 & 48.50 & 57.51 & 53.61 \\
JAT~\cite{li2025towards}   & 55.32 & 50.08 & 50.12 & 59.23 & 50.18 & 56.22 & 53.52 \\
MBCD~\cite{wang2025modality}   & \secnd{56.22} & \best{55.30} & \best{53.41} & 61.17 & 53.64 & \best{62.26} & \best{57.00} \\
GMP~\cite{li2026balancing}  & 53.17 & 49.82 & 48.97 & 59.65 & 49.81 & 57.33 & 53.12 \\
\textit{Oracle}   & \textit{76.13} & \textit{76.80} & \textit{60.23} & \textit{76.80} & \textit{60.23} & \textit{76.13} & \textit{71.05} \\

\bottomrule
\end{tabular}
\end{threeparttable}
}
\vspace{-0.5em}
\label{tab:epic-ssdg-va}
\end{table*}

\section{Robustness under Corruptions and Missing Modalities}
\label{sec:robustness}

Real-world deployments frequently expose multimodal systems to corrupted inputs and missing modalities, yet these critical scenarios remain largely underexplored in MMDG research. To evaluate robustness under realistic sensor failures, we adopt two representative corruptions commonly studied in the literature~\cite{dong2025aeo}: wind noise in the audio stream and defocus blur in the video stream. We further assess missing-modality generalization by removing either video or audio during inference.

\begin{figure}[t]
  \centering  \includegraphics[width=\linewidth]{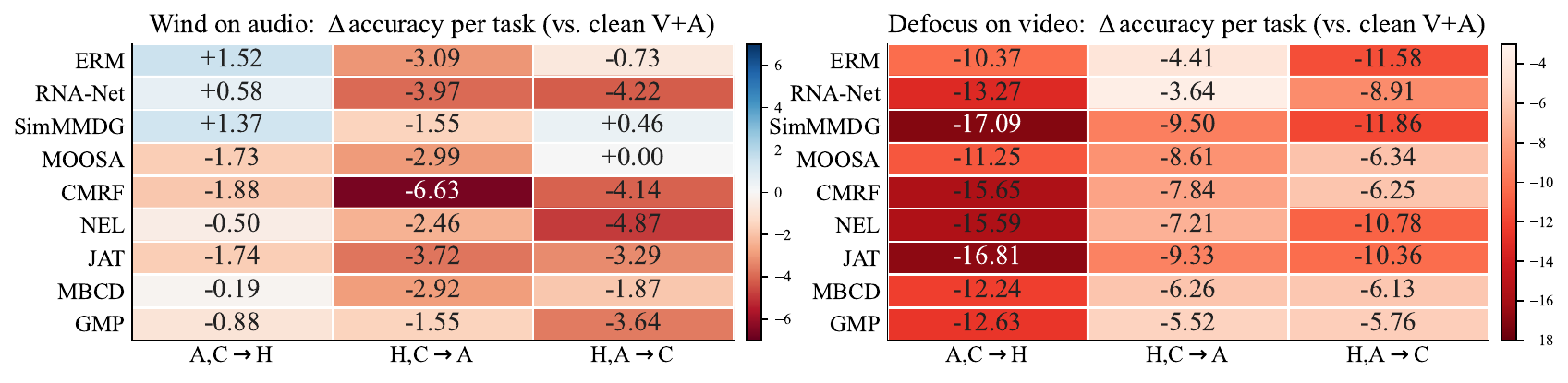}
  \vspace{-2em}
   \caption{Multimodal multi-source DG with \textbf{corruptions} on HAC dataset. Values show the change relative to the clean Video+Audio setting. Detailed results are in Table~\ref{tab:hac-dg-corruption-merged}. }
   \label{fig:hac-dg-corruptions}
\end{figure}

\textbf{Robustness under corruptions.}  Figure~\ref{fig:hac-dg-corruptions} reports multi-source DG performance on HAC under both corruptions, with subscripts indicating deviations from the clean V+A baseline. Under audio corruption, degradation is modest but widespread: all methods except SimMMDG decline by $0.77-4.22$ percentage points. 
Conversely, video corruption proves substantially more severe, causing accuracy drops of $7.97-12.82$ points. Crucially, performance rankings under corruption deviate markedly from clean-data rankings: MOOSA rises to first place, while SimMMDG drops from second to seventh. This rank inversion yields a critical takeaway: \textit{clean benchmark performance does not reliably predict deployment robustness under corruption}. This suggests that methods optimized for clean-domain alignment may overfit to modality-specific statistics, making them brittle when modality quality degrades. Notably, the most robust methods under defocus blur all incorporate explicit modality-balancing or competition-aware objectives, suggesting that these strategies inherently improve corruption robustness.

\begin{figure}[t]
  \centering  \includegraphics[width=\linewidth]{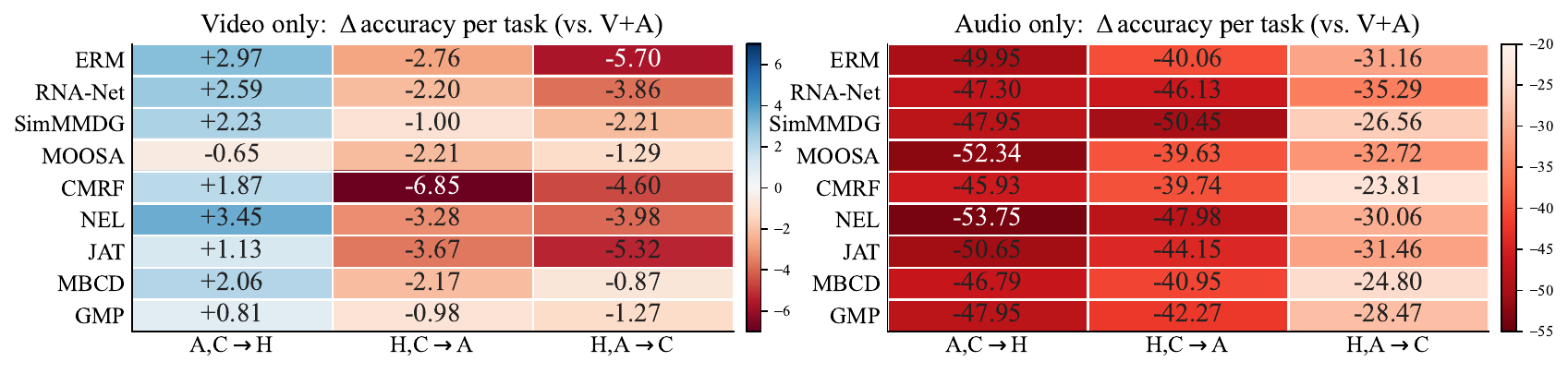}
  \vspace{-2em}
   \caption{Multimodal multi-source DG with \textbf{missing modalities} on HAC dataset. Values show the change relative to the full Video+Audio setting. Detailed results are in Table~\ref{tab:hac-dg-missing}. }
   \label{fig:hac-dg-missing}
\end{figure}

\textbf{Missing modalities.} Figure~\ref{fig:hac-dg-missing} evaluates robustness when a modality is unavailable at inference. We observe a striking asymmetry: removing audio causes only minor degradation ($0.32-3.20$ point drops), whereas removing video results in severe failures ($36.50-43.93$ point drops). For example, SimMMDG loses merely $0.33$ points when transitioning from V+A to V-only, but drops by $41.66$ points under A-only inference. Furthermore, in the A, C $\rightarrow$ H transfer setting, removing audio actually improves performance in most cases. This reveals a modality hierarchy under domain shift, where dominant modalities (e.g., video) govern robustness, while auxiliary modalities can introduce instability when not properly integrated. Appendix~\ref{sec:robust-beyond} extends this evaluation to fault diagnosis and sentiment analysis, and Appendix~\ref{sec:guidelines} shows that random modality dropout mitigates missing-modality failure.


\begin{tcolorbox}[takeaways]
\begin{enumerate}[leftmargin=*]
\setlength\itemsep{0em}
\item Clean benchmark performance does not reliably predict real-world robustness: methods that perform strongly under standard DG settings can degrade substantially under realistic input corruptions.
\item Video remains the dominant modality in action recognition, while audio often provides only auxiliary benefits and can even reduce performance when multimodal integration is not carefully optimized.
\item Current MMDG methods remain highly vulnerable to sensor failures, highlighting the urgent need for future approaches that explicitly address corruption robustness and missing-modality resilience.
\end{enumerate}
\end{tcolorbox}

\begin{table*}[t!]
\centering
\caption{Multimodal \textbf{misclassification detection} on HAC with video and audio modalities.}
\vspace{0.4em}
\setlength{\tabcolsep}{2.5pt}
\renewcommand{\arraystretch}{1.1}
\resizebox{0.95\linewidth}{!}{
\begin{threeparttable}
\begin{tabular}{l ccc ccc ccc ccc}  
\toprule
\textbf{Method} & \multicolumn{3}{c}{A, C $\rightarrow$ H} & \multicolumn{3}{c}{H, C $\rightarrow$ A} & \multicolumn{3}{c}{H, A $\rightarrow$ C}  & \multicolumn{3}{c}{\textit{Mean}} \\
\cmidrule(lr){2-4} \cmidrule(lr){5-7} \cmidrule(lr){8-10} \cmidrule(lr){11-13}
& AURC$\downarrow$ & AUROC$\uparrow$ & FPR95$\downarrow$ & AURC$\downarrow$ & AUROC$\uparrow$ & FPR95$\downarrow$ & AURC$\downarrow$ & AUROC$\uparrow$ & FPR95$\downarrow$& AURC$\downarrow$ & AUROC$\uparrow$ & FPR95$\downarrow$ \\
\midrule
ERM & 75.02 & 84.62 & 73.95 & 73.26 & 84.67 & 59.31 & 271.82 & 74.22 & 85.40 & 140.03 & 81.17 & 72.89 \\
RNA-Net~\cite{Planamente_2022_WACV} & 84.62 & 82.95 & 73.84 & 75.73 & 83.13 & 63.73 & 266.91 & 74.54 & 81.78 & 142.42 & 80.21 & 73.12 \\
SimMMDG~\cite{dong2023SimMMDG} & \best{58.94} & \best{86.06} & 68.01 & \secnd{67.61} & \best{85.19} & 68.84 & \best{237.51} & \best{76.42} & 84.20 & \best{121.35} & \best{82.56} & 73.68 \\
MOOSA~\cite{dong2024moosa} & 63.00 & 85.18 & \best{61.19} & \best{65.33} & 84.92 & \best{57.51} & \secnd{264.25} & 73.14 & 81.91 & \secnd{130.86} & 81.08 & \best{66.87} \\
CMRF~\cite{fan2024cross} & 69.36 & 85.89 & 66.88 & 83.75 & 81.93 & 77.21 & 359.01 & 69.07 & 86.43 & 170.71 & 78.96 & 76.84 \\
NEL~\cite{zhang2025nonpolarized} & 64.43 & 85.34 & 68.30 & 73.29 & 83.78 & 59.44 & 289.59 & \secnd{74.62} & \best{79.89} & 142.44 & 81.25 & \secnd{69.21} \\
JAT~\cite{li2025towards} & \secnd{62.33} & \secnd{85.96} & \secnd{63.65} & 68.46 & 84.47 & 66.37 & 268.85 & 74.18 & 84.22 & 133.21 & \secnd{81.54} & 71.41 \\
MBCD~\cite{wang2025modality} & 67.54 & 85.12 & 63.80 & 79.35 & 82.24 & 69.87 & 270.52 & 74.08 & 84.35 & 139.14 & 80.48 & 72.67 \\
GMP~\cite{li2026balancing} & 66.50 & 85.83 & 67.49 & 69.83 & \secnd{84.97} & \secnd{58.39} & 304.57 & 69.87 & 86.47 & 146.97 & 80.22 & 70.78 \\
\bottomrule
\end{tabular}
\end{threeparttable}
}
\vspace{-0.7em}
\label{tab:hac-misd}
\end{table*}

\section{Trustworthiness: Misclassification and Out-of-Distribution Detection}
\label{sec:trust}
 
Beyond classification accuracy, multimodal systems are also expected to identify when their predictions are likely to be incorrect (misclassification detection~\cite{liu2026cvpr}) and to detect inputs that are semantically novel (out-of-distribution detection~\cite{li2024dpu,liu2025fm}). This is the first standardized evaluation of trustworthiness in MMDG. We evaluate both capabilities on HAC using the V+A combination. For OOD detection, HAC serves as the in-distribution dataset, while EPIC-Kitchens is used as the OOD dataset. For misclassification detection (MisD), we report AURC (Area Under the Risk-Coverage Curve), AUROC, and FPR95 (false positive rate at 95\% true positive rate). For OOD detection, we report AUROC and FPR95.
 
\textbf{Misclassification detection.} Table~\ref{tab:hac-misd} presents the MisD results. SimMMDG achieves the strongest overall performance (best mean AURC and AUROC), suggesting that its explicit decomposition of modality-shared and modality-specific features yields better-calibrated uncertainty estimates. Meanwhile, MOOSA achieves the best mean FPR95, indicating that its self-supervised pretext tasks generate confidence scores that effectively separate correct from incorrect predictions. In contrast, while CMRF maintains competitive classification accuracy, it ranks last across all MisD metrics. This discrepancy exposes a critical disconnect between predictive accuracy and model trustworthiness, a vulnerability largely overlooked in prior MMDG research.
 
\textbf{Out-of-distribution detection.} Table~\ref{tab:hac-ood} reports the OOD detection results, where SimMMDG again achieves the strongest overall performance. Interestingly, CMRF, which ranks last in MisD, achieves the second-highest mean OOD AUROC. This confirms that these two trustworthiness dimensions are non-redundant: mechanisms that improve OOD separation can simultaneously degrade confidence calibration for in-distribution errors. The inverse pattern is also observed: MOOSA attains the best MisD FPR95 but falls to the bottom in OOD AUROC. Furthermore, despite its exceptional classification accuracy across EPIC-Kitchens and HAC, MBCD yields only moderately on OOD AUROC and MisD metrics. Ultimately, these findings demonstrate that high predictive accuracy does not guarantee model trustworthiness, and even trust-oriented metrics may favor different methods depending on whether the focus is on misclassification calibration or OOD detection. Appendix~\ref{sec:trust-beyond} confirms these trends on fault diagnosis.

\begin{tcolorbox}[takeaways]
\begin{enumerate}[leftmargin=*]
\setlength\itemsep{0em}
\item Predictive accuracy does not guarantee model trustworthiness. High classification performance does not inherently translate to reliable confidence estimation.
\item Misclassification detection and OOD detection are non-redundant capabilities. A model's ability to separate in-distribution from OOD samples does not predict its ability to properly calibrate confidence on in-distribution errors.
\item Current MMDG methods exhibit substantial limitations in uncertainty calibration, highlighting the need for future approaches that jointly optimize accuracy and trustworthiness.
\end{enumerate}
\end{tcolorbox}

\begin{table*}[t!]
\centering
\caption{Multimodal \textbf{out-of-distribution detection} with video and audio modalities, where HAC is the ID dataset and EPIC-Kitchens as OOD dataset.}
\vspace{0.4em}
\resizebox{0.9\linewidth}{!}{
\begin{threeparttable}
\begin{tabular}{l ccc ccc ccc c} 
\toprule
\textbf{Method} & \multicolumn{2}{c}{A, C $\rightarrow$ H} & \multicolumn{2}{c}{H, C $\rightarrow$ A} & \multicolumn{2}{c}{H, A $\rightarrow$ C}  & \multicolumn{2}{c}{\textit{Mean}} \\
\cmidrule(lr){2-3} \cmidrule(lr){4-5} \cmidrule(lr){6-7} \cmidrule(lr){8-9}
&  AUROC$\uparrow$ & FPR95$\downarrow$ &  AUROC$\uparrow$ & FPR95$\downarrow$ &  AUROC$\uparrow$ & FPR95$\downarrow$ &  AUROC$\uparrow$ & FPR95$\downarrow$ \\
\midrule
ERM & 70.63 & 62.00 & 53.64 & 88.08 & \best{46.05} & \best{90.53} & 56.77 & 80.20 \\
RNA-Net~\cite{Planamente_2022_WACV} & 68.21 & 67.63 & 57.68 & 83.77 & 38.56 & 97.61 & 54.82 & 83.00 \\
SimMMDG~\cite{dong2023SimMMDG} & \secnd{77.19} & \secnd{53.42} & \best{73.12} & \best{62.14} & 35.18 & 96.50 & \best{61.83} & \best{70.69} \\
MOOSA~\cite{dong2024moosa} & 67.23 & 75.13 & 64.65 & 69.43 & 34.29 & 98.25 & 55.39 & 80.94 \\
CMRF~\cite{fan2024cross} & \best{77.61} & \best{52.34} & 60.04 & 79.58 & 40.70 & 95.59 & \secnd{59.45} & 75.84 \\
NEL~\cite{zhang2025nonpolarized} & 69.32 & 65.81 & 63.20 & 78.62 & 37.21 & 96.65 & 56.58 & 80.36 \\
JAT~\cite{li2025towards} & 72.71 & 61.82 & 63.62 & 70.24 & \secnd{41.20} & \secnd{95.33} & 59.18 & 75.80 \\
MBCD~\cite{wang2025modality}  & 76.44 & 53.46 & \secnd{65.87} & \secnd{68.16} & 34.11 & 98.54 & 58.81 & \secnd{73.39} \\
GMP~\cite{li2026balancing} & 71.54 & 61.48 & 55.37 & 86.46 & 39.28 & 96.79 & 55.40 & 81.58 \\
\bottomrule
\end{tabular}
\end{threeparttable}
}
\vspace{-0.5em}
\label{tab:hac-ood}
\end{table*}

\section{Comparison with Foundation Models}
\label{sec:foundation}

The nine evaluated methods are trained MMDG approaches. To examine whether our conclusions extend to foundation models, we additionally evaluate three CLIP variants (RN50, ViT-B/16, and ViT-L/14)~\cite{clip} and two Qwen2.5-VL models (3B and 7B)~\cite{bai2025qwen25vl} on EPIC-Kitchens and HAC (Table~\ref{tab:foundation-baselines}). These models are applied zero-shot to each target domain without training on the source domains.

\textbf{Foundation models excel under visual-style shift.} On HAC, where the domain shift is primarily visual style, CLIP ViT-L/14 achieves $91.08\%$ mean accuracy, substantially outperforming the best trained MMDG method (MOOSA, $70.95\%$) and approaching the target-trained Oracle ($92.81\%$). Web-scale pretraining thus provides strong robustness when the target shift is well covered by broad visual-semantic knowledge.

\textbf{Foundation models fall short on fine-grained egocentric actions.} On EPIC-Kitchens, the best foundation model (Qwen2.5-VL 3B) reaches only $39.71\%$, compared with $59.78\%$ for ERM and $64.87\%$ for MBCD with V+A. Generic vision-language pretraining alone therefore does not remove the need for task-specific MMDG training on fine-grained egocentric data. Moreover, these models do not natively support several modalities in our benchmark, including optical flow, raw audio, and vibration signals, which underscores the value of diverse task families for which specialized MMDG methods remain necessary.

Overall, foundation models and MMDG methods are complementary: the former excel at visual-style shifts, while the latter remain better suited to fine-grained tasks and specialized sensing modalities. Extending foundation models to non-visual sensing is a promising direction for future work.

\begin{table*}[t!]
\centering
\caption{Comparison with \textbf{foundation models} on EPIC-Kitchens and HAC datasets (accuracy, \%). CLIP~\cite{clip} and Qwen2.5-VL~\cite{bai2025qwen25vl} are applied zero-shot to each target domain without training on the source domains. For reference, we list multi-source MMDG results with video and audio from Table~\ref{tab:epic-hac-dg}: ERM, the best specialized method on each dataset (MBCD on EPIC-Kitchens and MOOSA on HAC), and the target-trained Oracle.}
\vspace{0.4em}
\resizebox{\linewidth}{!}{
\begin{threeparttable}
\begin{tabular}{llcccccccc}
\toprule
\textbf{Model} & \textbf{Backbone} & \multicolumn{4}{c}{\textbf{EPIC-Kitchens dataset}} & \multicolumn{4}{c}{\textbf{HAC dataset}} \\
\cmidrule(lr){3-6} \cmidrule(lr){7-10}
& & D1 & D2 & D3 & \textit{Mean} & H & A & C & \textit{Mean} \\
\midrule
\multirow{3}{*}{CLIP} & RN50 & 16.78 & 21.60 & 24.74 & 21.04 & 78.99 & 84.66 & 69.27 & 77.64 \\
& ViT-B/16 & 26.21 & 32.27 & 31.01 & 29.83 & 91.30 & 92.05 & 72.94 & 85.43 \\
& ViT-L/14 & 30.34 & 34.13 & 34.39 & 32.95 & \best{93.84} & \best{95.45} & \best{83.94} & \best{91.08} \\
\midrule
\multirow{2}{*}{Qwen2.5-VL} & 3B & 39.54 & 38.13 & 41.47 & 39.71 & 77.89 & 75.00 & 71.10 & 74.66 \\
& 7B & 34.02 & 36.40 & 36.65 & 35.69 & 86.59 & 91.47 & 83.02 & 87.03 \\
\midrule
\multicolumn{10}{l}{\textit{Reference: multi-source MMDG with video and audio (Table~\ref{tab:epic-hac-dg})}} \\
ERM & SlowFast + ResNet-18 & 57.47 & 61.20 & 60.68 & 59.78 & 75.91 & 77.48 & 53.40 & 68.93 \\
Best MMDG (MBCD / MOOSA) & SlowFast + ResNet-18 & \best{59.38} & \best{69.60} & \best{65.63} & \best{64.87} & 79.38 & 78.70 & 54.78 & 70.95 \\
\textit{Oracle} & SlowFast + ResNet-18 & \textit{60.23} & \textit{76.13} & \textit{76.80} & \textit{71.05} & \textit{92.75} & \textit{97.16} & \textit{88.53} & \textit{92.81} \\
\bottomrule
\end{tabular}
\end{threeparttable}
}
\label{tab:foundation-baselines}
\end{table*}

\section{Conclusion}
\label{sec:conclusion}
We introduce MMDG-Bench, the first unified benchmark for multimodal domain generalization, providing standardized evaluations across six datasets, three task families, six modality configurations, and nine representative methods in both multi- and single-source settings. Beyond clean-domain accuracy, MMDG-Bench systematically assesses corruption robustness, missing-modality generalization, misclassification detection, and out-of-distribution detection to rigorously evaluate real-world deployment capability. Our large-scale study reveals five key findings: (1) under fair evaluation, specialized methods yield only marginal gains over strong baselines; (2) no single method consistently dominates across datasets, modalities, or task families; (3) a substantial gap relative to the target-trained Oracle confirms that MMDG is far from solved; (4) trimodal fusion does not reliably outperform the strongest bimodal configurations; and (5) all methods remain highly vulnerable to corruptions and missing modalities, with some degrading model trustworthiness despite clean accuracy gains. Collectively, these results demonstrate that evaluating clean cross-domain performance alone is insufficient. Future MMDG research must prioritize modality competition, corruption resilience, and trustworthy uncertainty estimation as first-class objectives. We further diagnose the causes of these failures and distill practitioner guidelines validated by prospective experiments (Appendices~\ref{sec:diagnosis} and~\ref{sec:guidelines}). We hope MMDG-Bench serves as a rigorous, reproducible foundation to drive the development of robust, deployment-ready multimodal systems.

\section*{Acknowledgments}

This study was supported by ELLIS Institute Finland and Tampere University. We acknowledge CSC-IT Center for Science, Finland, for providing access to the Roihu supercomputer. We also acknowledge the computational resources provided by the Aalto Science-IT project through the Triton cluster. The authors also acknowledge the support of "In-service diagnostics of the catenary/pantograph and wheelset axle systems through intelligent algorithms" (SENTINEL) project, supported by the ETH Mobility Initiative.

{\small
\bibliographystyle{plain}
\bibliography{egbib}
}

\newpage

\appendix

\section{Scope of the Evaluated Distribution Shifts and the Standardized Protocol}
\label{sec:scope}

\textbf{Distribution shifts covered.} MMDG-Bench targets four distinct and practically relevant forms of distribution shift.
\textit{(i) Environment shift} (EPIC-Kitchens): transfer across kitchens that differ in layout, lighting, objects, and wearers, reflecting common deployment settings in egocentric and robotic perception and following the established MM-SADA protocol~\cite{Munro_2020_CVPR}.
\textit{(ii) Style and appearance shift} (HAC): the same actions performed by humans, animals, and cartoon characters, providing a controlled probe of appearance variation analogous to the sketch and painting domains of PACS~\cite{li2017deeper} while largely preserving action semantics.
\textit{(iii) Operating-condition shift} (HUST Motor): transfer across rotational speeds from 5 to 30\,Hz, mirroring industrial fault diagnosis in which equipment is tested under operating conditions absent from training.
\textit{(iv) Cultural and language shift} (CMU-MOSI/CMU-MOSEI $\rightarrow$ CH-SIMS): cross-lingual and cross-cultural sentiment transfer, a realistic challenge for affective-computing systems deployed across languages and populations.
Together, these settings cover environmental, appearance, operating-condition, and cultural shifts across three task families. Our conclusions are grounded in these representative settings and are not intended to cover every possible multimodal distribution shift.

\textbf{What the standardized protocol controls.} Following DomainBed~\cite{gulrajani2020search}, all nine methods share, for each dataset: (i) identical data splits and preprocessing; (ii) identical backbones and feature extractors; (iii) identical optimizers, batch sizes, and training budgets; (iv) an identical hyperparameter protocol, consisting of the default configuration plus 10 random-search trials per algorithm-dataset pair drawn from the search spaces in Table~\ref{tab:hyper}; (v) model selection exclusively via training-domain validation, without any target-domain access; and (vi) retraining of the selected configuration with three random seeds, reporting the mean (and, in Appendix~\ref{sec:variance}, the standard deviation). All code, configurations, and splits are publicly released, so every design decision is inspectable and re-runnable.

\textbf{What it does not control.} Ten random-search trials cannot guarantee that each method reaches its global optimum, and shared backbones may favor some methods over others. These limitations, however, apply symmetrically to all methods, including ERM, and therefore cannot manufacture our central finding of marginal gains over ERM: under-tuned specialized methods and under-tuned ERM face the same handicap. Throughout the paper, ``fair comparison'' should thus be read as \emph{standardized evaluation under a uniform, fully documented protocol}.

\section{Statistical Significance and Ranking Stability}
\label{sec:variance}

All results in the main paper are averaged over three random seeds. To assess whether the small differences between methods are statistically meaningful, Table~\ref{tab:hac-dg-std} reports the mean and standard deviation across seeds for multi-source MMDG on HAC with video and audio, and Table~\ref{tab:hac-seed-rank} ranks all nine methods independently under each seed. The standard deviations capture variability arising from random initialization and stochastic optimization.

\textbf{Differences between leading methods lie within seed noise.} The three leading specialized methods, MOOSA ($70.95 \pm 1.26$), SimMMDG ($70.81 \pm 1.13$), and MBCD ($70.17 \pm 0.95$), differ by less than one standard deviation, so their relative order should not be interpreted as a reliable performance advantage. The gap between the best specialized method (MOOSA) and ERM ($68.93 \pm 1.59$) is $2.02$ percentage points, comparable to the observed standard deviations. This provides variance-aware support for our main finding that specialized MMDG methods offer only marginal gains over ERM under standardized evaluation.

\textbf{Rankings are stable at the extremes and noisy in the middle.} The per-seed rankings in Table~\ref{tab:hac-seed-rank} are consistent at both ends: SimMMDG and MOOSA always occupy the top two positions, MBCD ranks third or fourth, and NEL is always last, whereas the remaining methods frequently exchange positions across seeds. Together with the overlapping mean $\pm$ std intervals, this indicates that fine-grained rank differences among closely performing methods largely reflect random-seed noise and should not be over-interpreted.

\begin{table*}[t!]
\centering
\caption{Multimodal multi-source DG with video and audio on HAC dataset, reported as mean $\pm$ standard deviation (accuracy, \%) over three random seeds. Best and second-best mean results are in bold and underlined, respectively.}
\vspace{0.4em}
\setlength{\tabcolsep}{2.5pt}
\resizebox{0.65\linewidth}{!}{
\begin{threeparttable}
\begin{tabular}{lcccc}
\toprule
\textbf{Method} & A, C $\rightarrow$ H & H, C $\rightarrow$ A & H, A $\rightarrow$ C & \textit{Mean}\\
\midrule
ERM & $75.91_{\pm 1.78}$ & $77.48_{\pm 1.71}$ & $53.40_{\pm 1.29}$ & $68.93_{\pm 1.59}$ \\
RNA-Net~\cite{Planamente_2022_WACV} & $75.20_{\pm 1.89}$ & $77.48_{\pm 2.05}$ & $53.58_{\pm 1.21}$ & $68.75_{\pm 1.71}$ \\
SimMMDG~\cite{dong2023SimMMDG} & $\secnd{78.59}_{\pm 0.94}$ & $78.04_{\pm 1.21}$ & $\best{55.79}_{\pm 1.25}$ & $\secnd{70.81}_{\pm 1.13}$ \\
MOOSA~\cite{dong2024moosa} & $\best{79.38}_{\pm 1.55}$ & $\secnd{78.70}_{\pm 1.46}$ & $\secnd{54.78}_{\pm 0.80}$ & $\best{70.95}_{\pm 1.26}$ \\
CMRF~\cite{fan2024cross} & $77.94_{\pm 2.34}$ & $78.26_{\pm 1.16}$ & $51.84_{\pm 1.12}$ & $69.35_{\pm 1.54}$ \\
NEL~\cite{zhang2025nonpolarized} & $76.33_{\pm 2.20}$ & $76.42_{\pm 3.15}$ & $51.07_{\pm 1.85}$ & $67.94_{\pm 1.80}$ \\
JAT~\cite{li2025towards} & $78.16_{\pm 1.17}$ & $77.99_{\pm 1.88}$ & $53.11_{\pm 1.25}$ & $69.75_{\pm 0.87}$ \\
MBCD~\cite{wang2025modality} & $78.12_{\pm 1.03}$ & $\best{78.91}_{\pm 0.76}$ & $53.49_{\pm 2.37}$ & $70.17_{\pm 0.95}$ \\
GMP~\cite{li2026balancing} & $77.36_{\pm 1.78}$ & $76.47_{\pm 1.89}$ & $52.33_{\pm 3.33}$ & $68.72_{\pm 2.13}$ \\
\bottomrule
\end{tabular}
\end{threeparttable}
}
\label{tab:hac-dg-std}
\end{table*}

\begin{table}[t!]
\centering
\caption{Per-seed ranking (1 = best) of the nine methods for multimodal multi-source DG with video and audio on HAC dataset. Under each random seed, all methods are ranked independently by accuracy.}
\vspace{0.4em}
\setlength{\tabcolsep}{5pt}
\begin{tabular}{lccccccccc}
\toprule
 & ERM & RNA-Net & SimMMDG & MOOSA & CMRF & NEL & JAT & MBCD & GMP \\
\midrule
Seed 1 & 6 & 7 & 1 & 2 & 5 & 9 & 4 & 3 & 8 \\
Seed 2 & 5 & 8 & 1 & 2 & 3 & 9 & 6 & 4 & 7 \\
Seed 3 & 7 & 8 & 2 & 1 & 6 & 9 & 4 & 3 & 5 \\
\bottomrule
\end{tabular}
\label{tab:hac-seed-rank}
\end{table}

\section{Additional Analyses on Action Recognition}
\label{sec:add-ar}

This section complements Section~\ref{sec:fair} with five additional analyses on EPIC-Kitchens and HAC: transformer backbones, an encoder-controlled comparison of video and audio, foundation-model baselines, per-class accuracy, and an effective-robustness study that relates in-distribution to out-of-distribution accuracy.

\subsection{Results with Transformer Backbones}
\label{sec:transformer}

The main experiments use convolutional backbones (SlowFast for video and ResNet-18 for audio). To verify that our conclusions are not tied to this choice, we re-evaluate five representative methods on EPIC-Kitchens and HAC (V+A) with VideoMAE~\cite{tong2022videomae} as the video encoder and the Audio Spectrogram Transformer (AST)~\cite{gong2021ast} as the audio encoder (Table~\ref{tab:epic-hac-transformer}). On HAC, ERM reaches a mean accuracy of $74.73\%$, while the best method, JAT, reaches $76.23\%$, an improvement of $1.50$ percentage points; SimMMDG performs on par with ERM, whereas MOOSA and MBCD fall below it. On EPIC-Kitchens, all five methods lie within $1.8$ points of each other, and the best method (MBCD) outperforms ERM by only $0.69$ points. Specialized MMDG methods therefore offer limited and inconsistent improvements over ERM regardless of the backbone architecture.

\begin{table*}[t!]
\centering
\caption{Multimodal multi-source DG with video and audio on EPIC-Kitchens and HAC datasets using \textbf{transformer backbones}: VideoMAE~\cite{tong2022videomae} for video and the Audio Spectrogram Transformer (AST)~\cite{gong2021ast} for audio. Results are mean $\pm$ standard deviation (accuracy, \%) over three random seeds. Best and second-best mean results are in bold and underlined, respectively.}
\vspace{0.4em}
\setlength{\tabcolsep}{3.5pt}
\resizebox{\linewidth}{!}{
\begin{threeparttable}
\begin{tabular}{lcccccccc}
\toprule
& \multicolumn{4}{c}{\textbf{EPIC-Kitchens dataset}} & \multicolumn{4}{c}{\textbf{HAC dataset}} \\
\cmidrule(lr){2-5} \cmidrule(lr){6-9}
\textbf{Method} & D2, D3 $\rightarrow$ D1 & D1, D3 $\rightarrow$ D2 & D1, D2 $\rightarrow$ D3 & \textit{Mean} & A, C $\rightarrow$ H & H, C $\rightarrow$ A & H, A $\rightarrow$ C & \textit{Mean} \\
\midrule
ERM & $\secnd{51.18}_{\pm 1.66}$ & $58.71_{\pm 1.47}$ & $58.14_{\pm 1.15}$ & $56.01_{\pm 1.43}$ & $\best{89.42}_{\pm 1.12}$ & $83.46_{\pm 1.99}$ & $51.31_{\pm 1.71}$ & $74.73_{\pm 1.61}$ \\
SimMMDG~\cite{dong2023SimMMDG} & $49.65_{\pm 0.61}$ & $58.66_{\pm 2.20}$ & $\secnd{59.22}_{\pm 1.28}$ & $55.84_{\pm 1.36}$ & $88.68_{\pm 2.90}$ & $\secnd{85.43}_{\pm 0.98}$ & $50.50_{\pm 1.62}$ & $\secnd{74.87}_{\pm 1.83}$ \\
MOOSA~\cite{dong2024moosa} & $48.73_{\pm 1.61}$ & $56.53_{\pm 0.67}$ & $\best{59.55}_{\pm 0.89}$ & $54.94_{\pm 1.06}$ & $\secnd{89.33}_{\pm 1.49}$ & $81.94_{\pm 1.25}$ & $50.18_{\pm 3.43}$ & $73.81_{\pm 2.05}$ \\
JAT~\cite{li2025towards} & $50.80_{\pm 1.28}$ & $\best{59.73}_{\pm 1.27}$ & $57.63_{\pm 0.95}$ & $\secnd{56.05}_{\pm 1.17}$ & $88.39_{\pm 1.52}$ & $\best{86.08}_{\pm 1.42}$ & $\best{54.22}_{\pm 1.03}$ & $\best{76.23}_{\pm 1.32}$ \\
MBCD~\cite{wang2025modality} & $\best{51.28}_{\pm 0.26}$ & $\secnd{59.64}_{\pm 1.48}$ & $59.17_{\pm 1.28}$ & $\best{56.70}_{\pm 1.01}$ & $88.22_{\pm 1.38}$ & $79.75_{\pm 2.28}$ & $\secnd{52.42}_{\pm 2.81}$ & $73.46_{\pm 2.15}$ \\
\bottomrule
\end{tabular}
\end{threeparttable}
}
\label{tab:epic-hac-transformer}
\end{table*}

\subsection{Is Video Dominance an Encoder Artifact?}
\label{sec:encoder}

In Section~\ref{sec:robustness}, removing video hurts far more than removing audio. Because video uses a Kinetics-pretrained SlowFast backbone while audio uses a ResNet-18, this asymmetry could in principle reflect encoder strength rather than the information content of each modality. To disentangle the two, we replace ResNet-18 with the substantially stronger AST and compare unimodal multi-source DG on HAC (Table~\ref{tab:hac-unimodal}). Audio with AST reaches only $28.51\%$ mean accuracy, far below video with SlowFast ($66.75\%$) and close to audio-only inference with the original ResNet-18 (e.g., $28.54\%$ for ERM in Table~\ref{tab:hac-dg-missing}). Encoder capacity is thus unlikely to be the main bottleneck: the modality hierarchy primarily reflects differences in the discriminative and transferable information carried by each modality.

\begin{table}[t!]
\centering
\caption{Unimodal multi-source DG on HAC dataset (accuracy, \%) with a strong transformer audio encoder (AST~\cite{gong2021ast}) versus the video encoder used in the main experiments (SlowFast~\cite{Feichtenhofer_2019_ICCV}). Audio remains far behind video even with the stronger encoder.}
\vspace{0.4em}
\setlength{\tabcolsep}{6pt}
\begin{tabular}{llcccc}
\toprule
\textbf{Modality} & \textbf{Encoder} & A, C $\rightarrow$ H & H, C $\rightarrow$ A & H, A $\rightarrow$ C & \textit{Mean} \\
\midrule
Audio & AST & 29.70 & 32.78 & 23.06 & 28.51 \\
Video & SlowFast & 77.39 & 74.82 & 48.03 & 66.75 \\
\bottomrule
\end{tabular}
\label{tab:hac-unimodal}
\end{table}

\subsection{Per-class Accuracy}
\label{sec:perclass}

Top-1 accuracy alone cannot reveal whether a model collapses to majority classes on imbalanced data. Table~\ref{tab:hac-perclass} therefore reports per-class accuracy, together with overall top-1 accuracy, for all methods on each HAC target domain in the multi-source V+A setting. First, no method collapses: every method correctly recognizes samples from all seven classes in every target domain. Second, the per-class view exposes a failure mode hidden by top-1 accuracy. All methods struggle on \textit{drinking} when transferring to Animal ($9.0-15.8\%$) and Cartoon ($3.2-25.4\%$), compared with Human ($43.4-58.1\%$), likely because drinking is visually and acoustically similar to eating and varies substantially across styles. No specialized method consistently resolves this weakness relative to ERM, and most aggregate gains arise from easier classes.

\begin{table*}[t!]
\centering
\caption{\textbf{Per-class accuracy} (\%) for multimodal multi-source DG with video and audio on HAC dataset. For each target domain, we report the overall top-1 accuracy (\textit{Acc.}) and the accuracy on each of the seven classes (sleeping, watching TV, eating, drinking, swimming, running, and opening door). Best and second-best overall accuracies are in bold and underlined, respectively.}
\vspace{0.4em}
\setlength{\tabcolsep}{3.2pt}
\renewcommand{\arraystretch}{1.08}
\resizebox{0.8\linewidth}{!}{
\begin{threeparttable}
\begin{tabular}{lcccccccc}
\toprule
\multicolumn{9}{c}{\textbf{Target domain H: A, C $\rightarrow$ H}}\\
\midrule
\textbf{Method} & \textit{Acc.} & Sleep. & Watch. & Eat. & Drink. & Swim. & Run. & Door \\
\midrule
ERM     & 75.91 & 84.34 & 63.64 & 63.82 & 48.99 & 100.00 & 84.69 & 86.00 \\
RNA-Net~\cite{Planamente_2022_WACV} & 75.20 & 69.70 & 79.29 & 50.25 & 56.06 & 98.99 & 78.06 & 94.00 \\
SimMMDG~\cite{dong2023SimMMDG} & \secnd{78.59} & 77.27 & 87.88 & 68.34 & 43.94 & 100.00 & 83.16 & 89.50 \\
MOOSA~\cite{dong2024moosa}   & \best{79.38}  & 78.79 & 73.23 & 68.84 & 58.08 & 99.49 & 88.27 & 89.00 \\
CMRF~\cite{fan2024cross}    & 77.94 & 78.28 & 82.83 & 61.81 & 43.43 & 100.00 & 92.35 & 81.50 \\
NEL~\cite{zhang2025nonpolarized}     & 76.33 & 79.76 & 73.45 & 63.69 & 47.18 & 99.78 & 82.39 & 88.08 \\
JAT~\cite{li2025towards}     & 78.16 & 76.74 & 80.43 & 64.67 & 51.16 & 99.76 & 87.37 & 87.06 \\
MBCD~\cite{wang2025modality}    & 78.12 & 78.98 & 78.67 & 61.91 & 52.40 & 100.00 & 84.61 & 90.30 \\
GMP~\cite{li2026balancing}     & 77.36 & 75.76 & 79.45 & 60.69 & 53.18 & 100.00 & 86.39 & 86.14 \\
\midrule
\multicolumn{9}{c}{\textbf{Target domain A: H, C $\rightarrow$ A}}\\
\midrule
\textbf{Method} & \textit{Acc.} & Sleep. & Watch. & Eat. & Drink. & Swim. & Run. & Door \\
\midrule
ERM     & 77.48 & 76.98 & 89.51 & 80.41 & 13.53 & 98.50 & 95.00 & 94.44 \\
RNA-Net~\cite{Planamente_2022_WACV} & 77.48 & 82.01 & 82.52 & 89.86 & 9.02 & 95.49 & 95.00 & 93.33 \\
SimMMDG~\cite{dong2023SimMMDG} & 78.04 & 79.86 & 93.01 & 82.43 & 15.79 & 97.74 & 90.83 & 90.00 \\
MOOSA~\cite{dong2024moosa}   & \secnd{78.70} & 79.86 & 90.91 & 87.84 & 12.03 & 100.00 & 89.17 & 95.56 \\
CMRF~\cite{fan2024cross}    & 78.26 & 81.29 & 76.22 & 79.73 & 15.04 & 100.00 & 93.33 & 95.56 \\
NEL~\cite{zhang2025nonpolarized}     & 76.42 & 81.83 & 82.26 & 84.88 & 9.91  & 98.18 & 89.50 & 93.61 \\
JAT~\cite{li2025towards}     & 77.99 & 78.44 & 88.87 & 85.49 & 13.52 & 97.79 & 94.11 & 92.22 \\
MBCD~\cite{wang2025modality}    & \best{78.91} & 81.66 & 88.09 & 83.72 & 15.75 & 100.00 & 92.34 & 96.45 \\
GMP~\cite{li2026balancing}     & 76.47 & 76.53 & 86.96 & 80.58 & 14.61 & 98.88 & 92.20 & 90.31 \\
\midrule
\multicolumn{9}{c}{\textbf{Target domain C: H, A $\rightarrow$ C}}\\
\midrule
\textbf{Method} & \textit{Acc.} & Sleep. & Watch. & Eat. & Drink. & Swim. & Run. & Door \\
\midrule
ERM     & 53.40 & 65.09 & 68.60 & 39.26 & 11.90 & 54.73 & 75.70 & 50.95 \\
RNA-Net~\cite{Planamente_2022_WACV} & 53.58 & 56.60 & 42.98 & 33.74 & 25.40 & 77.70 & 69.63 & 57.14 \\
SimMMDG~\cite{dong2023SimMMDG} & \best{55.79} & 64.15 & 46.28 & 30.67 & 23.81 & 69.59 & 73.83 & 67.62 \\
MOOSA~\cite{dong2024moosa}   & \secnd{54.78} & 55.66 & 47.11 & 59.51 & 3.17 & 53.38 & 66.36 & 75.24 \\
CMRF~\cite{fan2024cross}    & 51.84 & 42.45 & 65.29 & 22.70 & 5.56 & 53.38 & 85.51 & 39.52 \\
NEL~\cite{zhang2025nonpolarized}     & 51.07 & 59.47 & 47.73 & 38.36 & 9.15  & 61.44 & 69.39 & 57.77 \\
JAT~\cite{li2025towards}     & 53.11 & 54.18 & 57.44 & 39.07 & 14.36 & 60.65 & 76.10 & 55.48 \\
MBCD~\cite{wang2025modality}    & 53.49 & 57.91 & 55.17 & 35.30 & 16.59 & 62.88 & 72.33 & 60.71 \\
GMP~\cite{li2026balancing}     & 52.33 & 52.94 & 56.20 & 33.33 & 17.62 & 63.91 & 74.86 & 54.24 \\
\bottomrule
\end{tabular}
\end{threeparttable}
}
\label{tab:hac-perclass}
\end{table*}

\section{Robustness and Trustworthiness Beyond Action Recognition}
\label{sec:beyond}

\subsection{Corruptions and Missing Modalities on Fault Diagnosis and Sentiment Analysis}
\label{sec:robust-beyond}

Section~\ref{sec:robustness} evaluates robustness on HAC with video and audio. To test whether the observed fragility is specific to one dataset and modality pair, we extend the evaluation to fault diagnosis (HUST Motor, vibration and acoustic, D1 $\rightarrow$ D2; Table~\ref{tab:hust-robust}) and sentiment analysis (CMU-MOSI $\rightarrow$ CH-SIMS, video, audio, and text; Table~\ref{tab:msa-robust}). In each setting, we either remove one modality at inference or inject Gaussian noise into the inputs, and report the change relative to the clean setting.

\textbf{Robustness failures generalize across tasks, with task-dependent severity.} On HUST, removing vibration or injecting Gaussian noise causes drops of up to $40.08$ and $46.17$ points, respectively; on MOSI $\rightarrow$ SIMS, removing audio or video reduces ACC2 by up to $30.20$ points. The fragility observed on HAC is therefore not specific to a single dataset, task, or modality pair. Its severity, however, depends strongly on the task: Gaussian noise is catastrophic for fault diagnosis (all methods fall to $16.58-27.64\%$) but costs at most $4.60$ ACC2 points in sentiment analysis.

\textbf{The dominant modality depends on the task.} Vibration is the critical modality for fault diagnosis: removing it hurts every method, whereas removing acoustic input is far less harmful for most methods and even improves ERM and RNA-Net. Conversely, on MOSI $\rightarrow$ SIMS, removing text improves or preserves ACC2 and improves F1 for all nine methods (up to $+3.33$ ACC2 and $+9.09$ F1), suggesting that text becomes unreliable under cross-lingual shift while audio and video carry the transferable signal. Conclusions about modality importance drawn from a single task therefore do not generalize.

\textbf{Clean and robust rankings differ, and no method is consistently robust.} On HUST, MBCD is the best clean method ($63.33\%$) and the most robust to missing vibration ($-10.25$), yet it degrades the most under missing acoustic input ($-47.17$) and Gaussian noise ($-46.17$), while RNA-Net shows the opposite trend. On MOSI $\rightarrow$ SIMS, MBCD and NEL degrade least under missing audio or video, whereas GMP, the third-best clean method, suffers the largest drop under missing audio ($-30.20$ ACC2). No method dominates across datasets or corruption types, reinforcing the conclusions of Section~\ref{sec:robustness}.

\begin{table*}[t!]
\centering
\caption{Multimodal single-source DG on HUST Motor dataset (D1 $\rightarrow$ D2) with vibration and acoustic modalities under \textbf{missing modalities} and \textbf{corruptions} (accuracy, \%). One modality is removed at inference, or Gaussian noise is injected into the inputs. Subscripts show the change relative to the clean setting (blue: improvement, red: degradation). Best and second-best results in each row are in bold and underlined, respectively.}
\vspace{0.4em}
\setlength{\tabcolsep}{2.5pt}
\resizebox{\linewidth}{!}{
\begin{threeparttable}
\begin{tabular}{lccccccccc}
\toprule
\textbf{Setting} & ERM & RNA-Net & SimMMDG & MOOSA & CMRF & NEL & JAT & MBCD & GMP \\
\midrule
Clean & 45.08 & 56.08 & 51.25 & 49.50 & 49.50 & 57.39 & 58.20 & \best{63.33} & \secnd{58.67} \\
Missing vibration & 20.08$_{\textcolor{red}{-25.00}}$ & 16.08$_{\textcolor{red}{-40.00}}$ & 29.08$_{\textcolor{red}{-22.17}}$ & 31.92$_{\textcolor{red}{-17.58}}$ & \secnd{33.00}$_{\textcolor{red}{-16.50}}$ & 18.14$_{\textcolor{red}{-39.25}}$ & 32.87$_{\textcolor{red}{-25.33}}$ & \best{53.08}$_{\textcolor{red}{-10.25}}$ & 18.59$_{\textcolor{red}{-40.08}}$ \\
Missing acoustic & 46.25$_{\textcolor{blue}{+1.17}}$ & \best{67.58}$_{\textcolor{blue}{+11.50}}$ & 44.50$_{\textcolor{red}{-6.75}}$ & 48.08$_{\textcolor{red}{-1.42}}$ & 43.58$_{\textcolor{red}{-5.92}}$ & 38.64$_{\textcolor{red}{-18.75}}$ & 43.70$_{\textcolor{red}{-14.50}}$ & 16.16$_{\textcolor{red}{-47.17}}$ & \secnd{55.84}$_{\textcolor{red}{-2.83}}$ \\
Gaussian noise & 16.83$_{\textcolor{red}{-28.25}}$ & \secnd{23.25}$_{\textcolor{red}{-32.83}}$ & 19.08$_{\textcolor{red}{-32.17}}$ & 16.58$_{\textcolor{red}{-32.92}}$ & 21.92$_{\textcolor{red}{-27.58}}$ & \best{27.64}$_{\textcolor{red}{-29.75}}$ & 17.95$_{\textcolor{red}{-40.25}}$ & 17.16$_{\textcolor{red}{-46.17}}$ & 18.75$_{\textcolor{red}{-39.92}}$ \\
\bottomrule
\end{tabular}
\end{threeparttable}
}
\label{tab:hust-robust}
\end{table*}

\begin{table*}[t!]
\centering
\caption{Multimodal single-source DG for \textbf{sentiment analysis} (CMU-MOSI $\rightarrow$ CH-SIMS) with video, audio, and text modalities under \textbf{missing modalities} and \textbf{corruptions}. One modality is removed at inference, or Gaussian noise is injected into the inputs. We report MAE (lower is better), ACC2, and F1 (higher is better). Subscripts show the change relative to the clean setting (blue: improvement, red: degradation).}
\vspace{0.4em}
\setlength{\tabcolsep}{2.5pt}
\resizebox{\linewidth}{!}{
\begin{threeparttable}
\begin{tabular}{ll*{9}{c}}
\toprule
\textbf{Metric} & \textbf{Setting} & ERM & RNA-Net & SimMMDG & MOOSA & CMRF & NEL & JAT & MBCD & GMP \\
\midrule
\multirow{5}{*}{MAE$\downarrow$}
& Clean
& 1.81 & 1.85 & 1.89 & 1.81 & 1.86 & 1.85 & 1.89 & 1.82 & 1.82 \\
& Missing text
& 1.77$_{\textcolor{blue}{-0.04}}$
& 1.81$_{\textcolor{blue}{-0.04}}$
& 1.91$_{\textcolor{red}{+0.02}}$
& 1.84$_{\textcolor{red}{+0.03}}$
& 1.89$_{\textcolor{red}{+0.03}}$
& 1.87$_{\textcolor{red}{+0.02}}$
& 1.92$_{\textcolor{red}{+0.03}}$
& 1.83$_{\textcolor{red}{+0.01}}$
& 1.81$_{\textcolor{blue}{-0.01}}$ \\
& Missing audio
& 1.87$_{\textcolor{red}{+0.06}}$
& 1.91$_{\textcolor{red}{+0.06}}$
& 1.90$_{\textcolor{red}{+0.01}}$
& 1.81$_{\textcolor{red}{+0.00}}$
& 1.87$_{\textcolor{red}{+0.01}}$
& 1.89$_{\textcolor{red}{+0.04}}$
& 1.97$_{\textcolor{red}{+0.08}}$
& 1.86$_{\textcolor{red}{+0.04}}$
& 1.95$_{\textcolor{red}{+0.13}}$ \\
& Missing video
& 1.87$_{\textcolor{red}{+0.06}}$
& 1.90$_{\textcolor{red}{+0.05}}$
& 1.91$_{\textcolor{red}{+0.02}}$
& 1.84$_{\textcolor{red}{+0.03}}$
& 1.88$_{\textcolor{red}{+0.02}}$
& 1.87$_{\textcolor{red}{+0.02}}$
& 1.95$_{\textcolor{red}{+0.06}}$
& 1.83$_{\textcolor{red}{+0.01}}$
& 1.87$_{\textcolor{red}{+0.05}}$ \\
& Gaussian noise
& 1.83$_{\textcolor{red}{+0.02}}$
& 1.86$_{\textcolor{red}{+0.01}}$
& 1.91$_{\textcolor{red}{+0.02}}$
& 1.82$_{\textcolor{red}{+0.01}}$
& 1.87$_{\textcolor{red}{+0.01}}$
& 1.85$_{\textcolor{red}{+0.00}}$
& 1.89$_{\textcolor{red}{+0.00}}$
& 1.82$_{\textcolor{red}{+0.00}}$
& 1.82$_{\textcolor{red}{+0.00}}$ \\
\midrule
\multirow{5}{*}{ACC2$\uparrow$}
& Clean
& 62.80 & 61.71 & 68.05 & 60.61 & 63.02 & 61.39 & 62.80 & 66.07 & 64.47 \\
& Missing text
& 65.86$_{\textcolor{blue}{+3.06}}$
& 64.34$_{\textcolor{blue}{+2.63}}$
& 69.84$_{\textcolor{blue}{+1.79}}$
& 63.30$_{\textcolor{blue}{+2.69}}$
& 66.35$_{\textcolor{blue}{+3.33}}$
& 61.83$_{\textcolor{blue}{+0.44}}$
& 66.08$_{\textcolor{blue}{+3.28}}$
& 66.07$_{\textcolor{blue}{+0.00}}$
& 64.91$_{\textcolor{blue}{+0.44}}$ \\
& Missing audio
& 45.73$_{\textcolor{red}{-17.07}}$
& 42.67$_{\textcolor{red}{-19.04}}$
& 46.17$_{\textcolor{red}{-21.88}}$
& 40.05$_{\textcolor{red}{-20.56}}$
& 50.59$_{\textcolor{red}{-12.43}}$
& 52.86$_{\textcolor{red}{-8.53}}$
& 48.36$_{\textcolor{red}{-14.44}}$
& 58.19$_{\textcolor{red}{-7.88}}$
& 34.27$_{\textcolor{red}{-30.20}}$ \\
& Missing video
& 44.86$_{\textcolor{red}{-17.94}}$
& 43.11$_{\textcolor{red}{-18.60}}$
& 56.42$_{\textcolor{red}{-11.63}}$
& 49.67$_{\textcolor{red}{-10.94}}$
& 54.27$_{\textcolor{red}{-8.75}}$
& 58.76$_{\textcolor{red}{-2.63}}$
& 52.52$_{\textcolor{red}{-10.28}}$
& 64.98$_{\textcolor{red}{-1.09}}$
& 54.40$_{\textcolor{red}{-10.07}}$ \\
& Gaussian noise
& 58.20$_{\textcolor{red}{-4.60}}$
& 57.55$_{\textcolor{red}{-4.16}}$
& 66.74$_{\textcolor{red}{-1.31}}$
& 59.52$_{\textcolor{red}{-1.09}}$
& 60.18$_{\textcolor{red}{-2.84}}$
& 60.51$_{\textcolor{red}{-0.88}}$
& 61.49$_{\textcolor{red}{-1.31}}$
& 65.19$_{\textcolor{red}{-0.88}}$
& 63.16$_{\textcolor{red}{-1.31}}$ \\
\midrule
\multirow{5}{*}{F1$\uparrow$}
& Clean
& 67.74 & 66.69 & 75.55 & 63.30 & 70.91 & 64.78 & 69.11 & 73.74 & 71.15 \\
& Missing text
& 74.65$_{\textcolor{blue}{+6.91}}$
& 72.67$_{\textcolor{blue}{+5.98}}$
& 79.62$_{\textcolor{blue}{+4.07}}$
& 67.33$_{\textcolor{blue}{+4.03}}$
& 74.94$_{\textcolor{blue}{+4.03}}$
& 67.29$_{\textcolor{blue}{+2.51}}$
& 78.20$_{\textcolor{blue}{+9.09}}$
& 75.26$_{\textcolor{blue}{+1.52}}$
& 73.66$_{\textcolor{blue}{+2.51}}$ \\
& Missing audio
& 40.14$_{\textcolor{red}{-27.60}}$
& 36.88$_{\textcolor{red}{-29.81}}$
& 46.81$_{\textcolor{red}{-28.74}}$
& 44.93$_{\textcolor{red}{-18.37}}$
& 50.30$_{\textcolor{red}{-20.61}}$
& 48.25$_{\textcolor{red}{-16.53}}$
& 46.16$_{\textcolor{red}{-22.95}}$
& 56.52$_{\textcolor{red}{-17.22}}$
& 32.24$_{\textcolor{red}{-38.91}}$ \\
& Missing video
& 39.11$_{\textcolor{red}{-28.63}}$
& 37.11$_{\textcolor{red}{-29.58}}$
& 52.90$_{\textcolor{red}{-22.65}}$
& 45.58$_{\textcolor{red}{-17.72}}$
& 53.73$_{\textcolor{red}{-17.18}}$
& 59.81$_{\textcolor{red}{-4.97}}$
& 51.83$_{\textcolor{red}{-17.28}}$
& 70.84$_{\textcolor{red}{-2.90}}$
& 51.56$_{\textcolor{red}{-19.59}}$ \\
& Gaussian noise
& 57.59$_{\textcolor{red}{-10.15}}$
& 57.62$_{\textcolor{red}{-9.07}}$
& 74.51$_{\textcolor{red}{-1.04}}$
& 62.08$_{\textcolor{red}{-1.22}}$
& 69.36$_{\textcolor{red}{-1.55}}$
& 63.52$_{\textcolor{red}{-1.26}}$
& 67.81$_{\textcolor{red}{-1.30}}$
& 73.55$_{\textcolor{red}{-0.19}}$
& 69.93$_{\textcolor{red}{-1.22}}$ \\
\bottomrule
\end{tabular}
\end{threeparttable}
}
\label{tab:msa-robust}
\end{table*}

\subsection{Misclassification and OOD Detection on Fault Diagnosis}
\label{sec:trust-beyond}

Section~\ref{sec:trust} evaluates trustworthiness on HAC with EPIC-Kitchens as OOD data. We extend both misclassification detection (MisD) and OOD detection to HUST Motor under the D1 $\rightarrow$ D2 shift with vibration and acoustic modalities (Table~\ref{tab:hust-trust}), a setting that differs from HAC in task, modality, data type, and domain. For OOD detection, the first four of the six health states serve as in-distribution classes and the remaining two as OOD classes. We do not include sentiment analysis because it is formulated as regression with continuous labels, for which classification-based MisD and OOD metrics are not directly defined.

\textbf{The accuracy-trustworthiness disconnect extends beyond action recognition.} On HUST, SimMMDG achieves the best OOD detection (AUROC $81.38\%$, FPR95 $42.50\%$) but the worst MisD performance (AUROC $60.18\%$, FPR95 $90.00\%$), confirming that strong OOD separation does not imply reliable detection of a model's own errors. Method rankings also change substantially across tasks: SimMMDG is the best MisD method on HAC but the last on HUST, whereas MBCD, only mid-ranked on HAC, becomes the strongest MisD method on HUST by a wide margin (AURC $134.13$ vs. $217.57$ for the runner-up). Most methods exhibit MisD FPR95 values near or above $80\%$, indicating that confidence estimation under domain shift remains challenging in industrial sensing. No method thus provides universally reliable confidence estimates across task families, while the non-redundancy of MisD and OOD detection holds across tasks.

\begin{table}[t!]
\centering
\caption{Multimodal \textbf{misclassification detection} (MisD) and \textbf{out-of-distribution detection} (OOD) on HUST Motor dataset (D1 $\rightarrow$ D2) with vibration and acoustic modalities. For OOD detection, the first four health states serve as in-distribution classes and the remaining two as OOD classes. Best and second-best results are in bold and underlined, respectively.}
\vspace{0.4em}
\setlength{\tabcolsep}{6pt}
\renewcommand{\arraystretch}{1.05}
\begin{tabular}{lccccc}
\toprule
& \multicolumn{3}{c}{\textbf{Misclassification detection}} & \multicolumn{2}{c}{\textbf{OOD detection}} \\
\cmidrule(lr){2-4} \cmidrule(lr){5-6}
\textbf{Method} & AURC$\downarrow$ & AUROC$\uparrow$ & FPR95$\downarrow$ & AUROC$\uparrow$ & FPR95$\downarrow$ \\
\midrule
ERM & 357.79 & 65.33 & 87.72 & 68.78 & 69.88 \\
RNA-Net~\cite{Planamente_2022_WACV} & 261.92 & 69.50 & 87.12 & 73.56 & 67.25 \\
SimMMDG~\cite{dong2023SimMMDG} & 398.57 & 60.18 & 90.00 & \best{81.38} & \best{42.50} \\
MOOSA~\cite{dong2024moosa} & 342.99 & 67.28 & 86.78 & 63.06 & \secnd{54.00} \\
CMRF~\cite{fan2024cross} & 303.42 & 71.34 & 84.35 & 69.28 & 76.50 \\
NEL~\cite{zhang2025nonpolarized} & 231.57 & \secnd{76.55} & \secnd{79.02} & 61.32 & 77.38 \\
JAT~\cite{li2025towards} & 263.81 & 70.81 & 82.51 & 60.77 & 73.75 \\
MBCD~\cite{wang2025modality} & \best{134.13} & \best{84.61} & \best{68.20} & 73.23 & 58.13 \\
GMP~\cite{li2026balancing} & \secnd{217.57} & 75.48 & 80.04 & \secnd{77.87} & 63.50 \\
\bottomrule
\end{tabular}
\label{tab:hust-trust}
\end{table}

\section{Why Do Current Methods Fail? A Diagnostic Summary}
\label{sec:diagnosis}

Combining the results in the main paper with the additional analyses above, we identify four factors that jointly explain why specialized MMDG methods improve only marginally over ERM and remain fragile under realistic conditions.

\textbf{Modality dominance reflects unequal reliability across domains, not insufficient encoder capacity.} Replacing the audio encoder with the substantially stronger AST does not close the gap to video on HAC (Appendix~\ref{sec:encoder}). Consistently, removing video causes $36.50-43.93$ point drops, whereas removing audio costs at most $3.20$ points and can even improve performance in the A, C $\rightarrow$ H setting (Section~\ref{sec:robustness}). Similar asymmetries arise on HUST and in sentiment transfer (Appendix~\ref{sec:robust-beyond}). Current fusion mechanisms therefore cannot reliably suppress an auxiliary modality when it becomes weak or misleading under domain shift.

\textbf{The inconsistent benefit of additional modalities points to unresolved optimization conflict.} Most methods fail to outperform their strongest bimodal variant with trimodal input, while the methods that explicitly address modality competition (CMRF, MBCD, and GMP) obtain more consistent but still modest gains of $1.36$, $0.59$, and $0.92$ points on HAC (Section~\ref{sec:fair}). Existing balancing mechanisms alleviate modality competition only partially and do not yet yield robust multimodal fusion under shift.

\textbf{Learned representations remain fragile for the classes most sensitive to domain shift.} Method rankings change substantially under corruption, indicating reliance on clean source statistics that do not remain stable when modality quality changes. The per-class analysis (Appendix~\ref{sec:perclass}) further shows that all methods perform similarly poorly on the most transfer-sensitive class, with only $3-25\%$ accuracy for drinking on the Animal and Cartoon targets, while most gains arise from easier classes.

\textbf{Confidence reliability is decoupled from predictive accuracy.} Accuracy, MisD, and OOD detection yield markedly different method rankings on both HAC and HUST (Section~\ref{sec:trust} and Appendix~\ref{sec:trust-beyond}). Methods that perform well in OOD detection can still perform poorly at identifying their own errors. Since none of the evaluated methods explicitly optimizes uncertainty under domain shift, confidence remains unreliable after distribution change.

Overall, the limited progress in MMDG stems from the interaction of modality dominance, incomplete conflict-aware fusion, fragile representations under shift, and unreliable uncertainty estimation, rather than from a single isolated factor.

\section{Practitioner Guidelines}
\label{sec:guidelines}

Beyond diagnosing why current methods fall short, MMDG-Bench supports concrete recommendations for practitioners facing domain shift today. Each guideline is grounded in quantitative evidence from the benchmark. The first two are validated prospectively: the hypothesis was derived from our diagnostics first and then tested in a new experiment.

\textbf{Guideline 1: Use textual information from foundation models as semantic anchors for representation learning.} Decomposition-based methods such as SimMMDG and MOOSA split each modality's representation into modality-shared and modality-specific components, but enforce this split only geometrically (e.g., by maximizing the $\ell_2$ distance between the two components), which does not guarantee that the modality-specific features capture meaningful modality-specific semantics. We observe that the CLIP text space already encodes this structure: prompts describing the same action through different modalities (e.g., ``a gentle audio of someone sleeping peacefully'', ``a video of a baby sleeping in a crib'', and ``optical flow of slow and calm sleep motions'') form clearly separated clusters, while modality-agnostic descriptions (e.g., ``a person sleeping peacefully in bed'') form a distinct cluster of their own. Text embeddings can thus serve as semantic anchors for disentanglement. Using GPT-generated prompt sets per action and per modality, we align the modality-shared features with the modality-agnostic prompts and each modality-specific feature with its modality-explicit prompts. This simple \emph{textual guidance} improves SimMMDG on HAC in all four modality configurations, by $0.71-2.94$ points (Table~\ref{tab:hac-text}), and the resulting modality-specific features become interpretable by construction, as each is tied to a human-readable description. \textit{Takeaway: purely geometric disentanglement objectives can be replaced or augmented with alignment to modality-conditioned text embeddings, a plug-in change that requires no architectural modification. More broadly, leveraging textual information and foundation models is a promising direction for improving multimodal generalization.}

\begin{table}[t!]
\centering
\caption{Multimodal multi-source DG on HAC dataset (mean accuracy over the three target domains, \%) for SimMMDG with and without \textbf{textual guidance}, which aligns modality-shared and modality-specific features with modality-agnostic and modality-specific text prompts, respectively. V: video, A: audio, F: optical flow. Subscripts show the change relative to SimMMDG.}
\vspace{0.4em}
\setlength{\tabcolsep}{6pt}
\begin{tabular}{lcccc}
\toprule
\textbf{Method} & V+A & V+F & A+F & V+A+F \\
\midrule
SimMMDG~\cite{dong2023SimMMDG} & 70.81 & 69.83 & 56.75 & 69.38 \\
SimMMDG + Textual Guidance & \best{71.52}$_{\textcolor{blue}{+0.71}}$ & \best{71.97}$_{\textcolor{blue}{+2.14}}$ & \best{59.69}$_{\textcolor{blue}{+2.94}}$ & \best{70.98}$_{\textcolor{blue}{+1.60}}$ \\
\bottomrule
\end{tabular}
\label{tab:hac-text}
\end{table}

\textbf{Guideline 2: Train with random modality dropout when missing modalities are expected at deployment.} Multi-source training can induce source-specific reliance on the modality that is most predictive during training, which transfers poorly when that modality is unavailable in the target domain. On HAC, removing audio at inference reduces the mean accuracy of ERM by only $1.83$ points, whereas removing video causes a $40.39$-point drop. We therefore train ERM with random modality dropout across the source domains (RMD-ERM), randomly masking one modality while retaining full-modality training samples. As shown in Table~\ref{tab:hac-rmd}, this leaves full-modality accuracy essentially unchanged ($68.93\%$ vs. $68.88\%$), while improving the mean accuracy by $1.07$ points with video only ($67.10\% \rightarrow 68.17\%$) and by $4.14$ points with audio only ($28.54\% \rightarrow 32.68\%$); when video is missing, RMD-ERM improves all three transfers. Although the large gap between modalities remains, the intervention consistently reduces missing-modality failure under domain shift. \textit{Takeaway: random modality dropout during multi-source training improves missing-modality generalization on unseen domains while preserving full-modality accuracy; it requires no target-domain data and no inference-time modification.}

\begin{table*}[t!]
\centering
\caption{Multimodal multi-source DG with \textbf{missing modalities} on HAC dataset (accuracy, \%) for ERM and ERM trained with \textbf{random modality dropout} (RMD-ERM). Subscripts show the change relative to the corresponding full Video+Audio model (blue: improvement, red: degradation).}
\vspace{0.4em}
\resizebox{0.8\linewidth}{!}{
\begin{threeparttable}
\begin{tabular}{lcccccc}
\toprule
& \multicolumn{2}{c}{\textbf{Modality}} & \multicolumn{4}{c}{\textbf{HAC dataset}}\\
\cmidrule(lr){2-3} \cmidrule(lr){4-7}
\textbf{Method} & Video & Audio & A, C $\rightarrow$ H & H, C $\rightarrow$ A & H, A $\rightarrow$ C & \textit{Mean}\\
\midrule
ERM & $\checkmark$ & $\checkmark$ & $75.91_{\pm 1.78}$ & $77.48_{\pm 1.71}$ & $53.40_{\pm 1.29}$ & $68.93_{\pm 1.59}$ \\
RMD-ERM & $\checkmark$ & $\checkmark$ & $76.59_{\pm 1.94}$ & $78.36_{\pm 1.52}$ & $51.68_{\pm 1.36}$ & $68.88_{\pm 1.61}$ \\
\midrule
ERM & $\checkmark$ & \ding{55} & 78.88$_{\textcolor{blue}{+2.97}}$ & 74.72$_{\textcolor{red}{-2.76}}$ & 47.70$_{\textcolor{red}{-5.70}}$ & 67.10$_{\textcolor{red}{-1.83}}$ \\
RMD-ERM & $\checkmark$ & \ding{55} & 78.56$_{\textcolor{blue}{+1.97}}$ & \best{75.89}$_{\textcolor{red}{-2.47}}$ & \best{50.06}$_{\textcolor{red}{-1.62}}$ & \best{68.17}$_{\textcolor{red}{-0.71}}$ \\
\midrule
ERM & \ding{55} & $\checkmark$ & 25.96$_{\textcolor{red}{-49.95}}$ & 37.42$_{\textcolor{red}{-40.06}}$ & 22.24$_{\textcolor{red}{-31.16}}$ & 28.54$_{\textcolor{red}{-40.39}}$ \\
RMD-ERM & \ding{55} & $\checkmark$ & \best{31.14}$_{\textcolor{red}{-45.45}}$ & \best{39.99}$_{\textcolor{red}{-38.37}}$ & \best{26.90}$_{\textcolor{red}{-24.78}}$ & \best{32.68}$_{\textcolor{red}{-36.20}}$ \\
\bottomrule
\end{tabular}
\end{threeparttable}
}
\label{tab:hac-rmd}
\end{table*}

\textbf{Guideline 3: Prioritize source-domain diversity over algorithm choice.} Our single- and multi-source results allow a direct comparison of two interventions available to a practitioner: collecting an additional diverse source domain versus adopting a specialized MMDG method. The former consistently dominates. On EPIC-Kitchens (V+A), moving ERM from single-source to multi-source training yields $+8.53$ points ($51.25\% \rightarrow 59.78\%$; Tables~\ref{tab:epic-ssdg-va} and~\ref{tab:epic-hac-dg}), whereas replacing ERM with the strongest specialized method within the multi-source setting yields only $+5.09$ points (MBCD, $64.87\%$), with most methods gaining $1-4$ points. On HAC (V+A), source diversity yields $+7.39$ points for ERM ($61.54\% \rightarrow 68.93\%$) versus $+2.02$ points for the best method swap (MOOSA, $70.95\%$). The effect is most dramatic on HUST Motor, where single-source training collapses (ERM mean $42.63\%$, with severe transfers as low as $1.75-18.14\%$; Table~\ref{tab:hust-ssdg}) while multi-source ERM reaches $69.90\%$, a $+27.27$-point gain from data diversity that is more than three times the largest algorithmic gain ($+8.33$ points, MOOSA). \textit{Takeaway: under domain shift today, one additional diverse source domain reliably yields more generalization than any specialized MMDG method in our benchmark.}

\textbf{Guideline 4: For trimodal fusion, use competition-aware methods; otherwise default to the strongest bimodal configuration.} Our trimodal results (Table~\ref{tab:epic-hac-dg}) indicate that modality competition is the primary bottleneck when adding modalities. On HAC, the three methods that explicitly handle modality competition gain consistently from trimodal fusion over their best bimodal configuration (CMRF $+1.36$, MBCD $+0.59$, GMP $+0.92$), whereas ERM, RNA-Net, SimMMDG, and MOOSA see minimal benefit or outright degradation (e.g., MOOSA declines from $71.29\%$ to $70.17\%$). \textit{Takeaway: if trimodal fusion is required, adopt a competition-aware method such as CMRF, MBCD, or GMP; otherwise, adding modalities without competition handling can hurt generalization, and the strongest bimodal configuration is a safer default.}

Together, these guidelines close the loop between diagnosis and action: the benchmark not only identifies why current methods fail, but also translates these insights into validated, readily deployable interventions.

\section{Related Work}
\label{sec:related}
 
\subsection{Domain Generalization}
 
Domain generalization (DG), formalized by~\cite{blanchard2011generalizing} and named by~\cite{muandet2013domain}, aims to learn models that transfer to unseen target distributions using only labeled source data, without target access during training. Comprehensive surveys~\cite{zhou2022domain,wang2022generalizing} categorize prior methodologies into four broad families. Domain alignment reduces source-domain feature divergence via moment matching~\cite{sun2016deep}, adversarial learning~\cite{ganin2015unsupervised}, or invariant risk minimization~\cite{arjovsky2019invariant}, positing that source-invariant representations will generalize to unseen targets. Meta-learning simulates domain shift by partitioning sources into pseudo-train and pseudo-test sets to optimize held-out performance~\cite{li2018learning}. Data augmentation~\cite{volpi2018generalizing} diversifies the training distribution through adversarial examples, mixup, or generative perturbations to cover potential test-domain shifts. Finally, regularization enforces solution properties conducive to out-of-distribution generalization, such as cross-domain gradient agreement~\cite{krueger2021out} or worst-case group robustness~\cite{sagawa2020distributionally}. Despite this methodological diversity, \cite{gulrajani2020search} demonstrated that under standardized evaluation, a carefully tuned ERM baseline matches or outperforms prominent DG algorithms across multiple benchmarks. This pivotal finding recentered the field on evaluation rigor, directly motivating our parallel investigation in the multimodal setting.
 
\subsection{Multimodal Domain Generalization}
 
Multimodal domain generalization (MMDG) extends DG to inputs comprising heterogeneous modalities (e.g., video, audio, text)~\cite{gungor2025integrating,huang2025bridging,chen2026towards,ji2026alignment}. This setting is uniquely challenging because modalities exhibit distinct statistical properties, converge at varying rates~\cite{wang2025modality}, and establish spurious cross-modal correlations that fracture under distribution shift~\cite{fan2024cross}. The canonical protocol originated with MM-SADA~\cite{Munro_2020_CVPR}, which defined the cross-kitchen action recognition task for domain adaptation, establishing the de facto MMDG benchmark. For DG specifically, RNA-Net~\cite{Planamente_2022_WACV} introduced Relative Norm Alignment to rebalance audio-visual feature norms across source domains. SimMMDG~\cite{dong2023SimMMDG} subsequently decomposed representations into modality-shared and -specific components, concurrently introducing the Human-Animal-Cartoon (HAC) dataset to stress-test cross-style generalization. MOOSA~\cite{dong2024moosa} later extended this approach to open-set MMDG via self-supervised pretext tasks.

More recent methods have targeted increasingly specific bottlenecks in the multimodal optimization landscape: CMRF~\cite{fan2024cross} flattens cross-modal representation spaces to address discrepant modality sharpness; NEL~\cite{zhang2025nonpolarized} mitigates embedding polarization; JAT~\cite{li2025towards} jointly applies adversarial training across modality and domain axes; MBCD~\cite{wang2025modality} replaces weight averaging with collaborative distillation and adaptive modality dropout; and GMP~\cite{li2026balancing} modulates gradients to resolve cross-modal conflicts. While adjacent domains like semi-supervised MMDG~\cite{li2026towards} and comprehensive surveys unifying multimodal adaptation and foundation models~\cite{dong2025mmdasurvey} have recently emerged, no prior work systematically consolidates MMDG evaluation across diverse task families, modality configurations, and robustness axes. MMDG-Bench directly addresses this critical gap.
 
\subsection{Domain Generalization Benchmarks}
 
The maturation of the DG field has been largely driven by community benchmarks. DomainBed~\cite{gulrajani2020search} standardized the evaluation of 14 algorithms across seven image datasets, revealing that prior reported gains largely stemmed from inconsistent evaluation protocols rather than algorithmic innovation. WILDS~\cite{koh2021wilds} extended benchmarking to 10 real-world datasets (e.g., satellite imagery, histopathology), demonstrating that substantial performance gaps persist on natural distribution shifts even for methods excelling on synthetic tasks. NICO++~\cite{zhang2023nico} introduced quantitative metrics for covariate and concept shift, showing that prior datasets occupied a narrow shift spectrum, and released a 200,000-image benchmark to expand this scope. Additionally, benchmarks like ImageNet-C~\cite{hendrycks2019benchmarking} and ImageNet-R~\cite{hendrycks2021many} have emerged to target specific failure modes, such as visual corruptions.

MMDG-Bench serves as the multimodal analogue to these foundational efforts. It provides a consolidated testbed that rigorously standardizes backbones, data splits, hyperparameters, and model selection across nine MMDG methods, six modality combinations, and six datasets. Furthermore, it introduces systematic evaluation axes for corruption robustness, missing modalities, and trustworthiness, critical dimensions completely absent from prior MMDG evaluations.

\section{Limitations, Broader Impacts, and Future Work}
\label{sec:future}

\subsection{Limitations} 
MMDG-Bench currently focuses on discriminative and regression tasks and does not yet cover other important settings such as multimodal retrieval or generative modeling. Additionally, our robustness evaluation is limited to two representative perturbations; extending this to broader, modality-specific corruption suites and adversarial attacks remains an important direction for future work. 

\subsection{Broader Impacts}
\textbf{Promoting Safe and Reliable AI Deployment:} By systematically exposing the vulnerabilities of current multimodal models to real-world noise, missing modalities, and out-of-distribution data, MMDG-Bench incentivizes the development of much safer AI systems. This is particularly crucial for high-stakes domains, such as industrial safety and predictive maintenance, where model failures can lead to physical harm or severe economic loss.

\textbf{Enhancing Model Transparency and Trust:} Our findings emphasize that high predictive accuracy does not guarantee reliable confidence estimation. By evaluating misclassification and out-of-distribution detection, our benchmark encourages the community to build AI systems that "know what they do not know." This transparency is essential for fostering meaningful human-AI collaboration and trust.

\subsection{Future Work}
 
Based on the comprehensive evaluations and findings from MMDG-Bench, it is evident that MMDG remains far from a solved problem. We identify several critical directions for future research to address the limitations of current approaches:

\textbf{Developing Beyond-Marginal Algorithms:} Current specialized MMDG methods offer only marginal improvements over strong baselines like ERM and fail to consistently dominate across diverse datasets or task families. Furthermore, a substantial gap to upper-bound performance persists. Future work must focus on discovering novel training paradigms or architectural innovations that genuinely generalize across task families, rather than overfitting to specific modality combinations or datasets.

\textbf{Addressing Modality Competition and Adaptive Fusion:} Our findings indicate that simply adding more modalities, such as through trimodal fusion, produces inconsistent benefits. In tasks like action recognition, dominant modalities (e.g., video) often overshadow auxiliary modalities (e.g., audio), which can even reduce performance when not properly integrated. These results highlight the need for dynamic and adaptive fusion mechanisms that explicitly address modality competition and optimally balance modality contributions based on context.

\textbf{Building Resilience to Real-World Corruptions and Sensor Failures:} Clean benchmark performance has proven to be a poor predictor of real-world robustness. Existing methods degrade substantially under realistic input corruptions and exhibit high vulnerability to missing modalities (sensor failures). Future MMDG frameworks must explicitly incorporate corruption robustness and missing-modality resilience into their optimization objectives, moving beyond idealized training environments.

\textbf{Jointly Optimizing Accuracy and Trustworthiness:} High predictive accuracy does not inherently translate to reliable confidence estimation. We observed that current models struggle with uncertainty calibration, and that misclassification detection and OOD detection represent non-redundant challenges. Future research should prioritize trustworthy MMDG by jointly optimizing for predictive accuracy and robust uncertainty quantification, ensuring models are safe and reliable in open-world deployments.

\section{Introduction of Datasets}
\label{sec:dataset}

We provide detailed information on the datasets included in MMDG-Bench, including action recognition, mechanical fault diagnosis, and sentiment analysis.

\subsection{Action Recognition}

\begin{figure}[t]
  \centering  \includegraphics[width=0.8\linewidth]{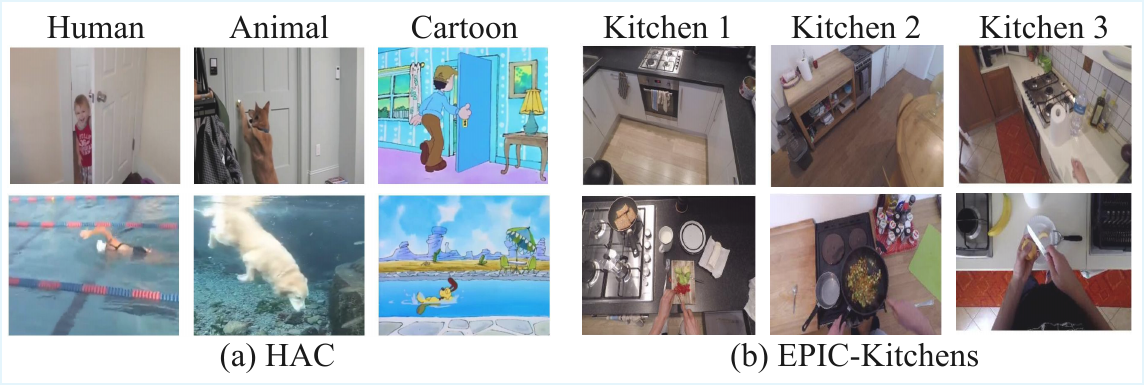}
  \vspace{-1em}
   \caption{Examples from action recognition datasets. }
   \label{fig:ar}
\end{figure}

\begin{figure}[t]
  \centering  \includegraphics[width=0.85\linewidth]{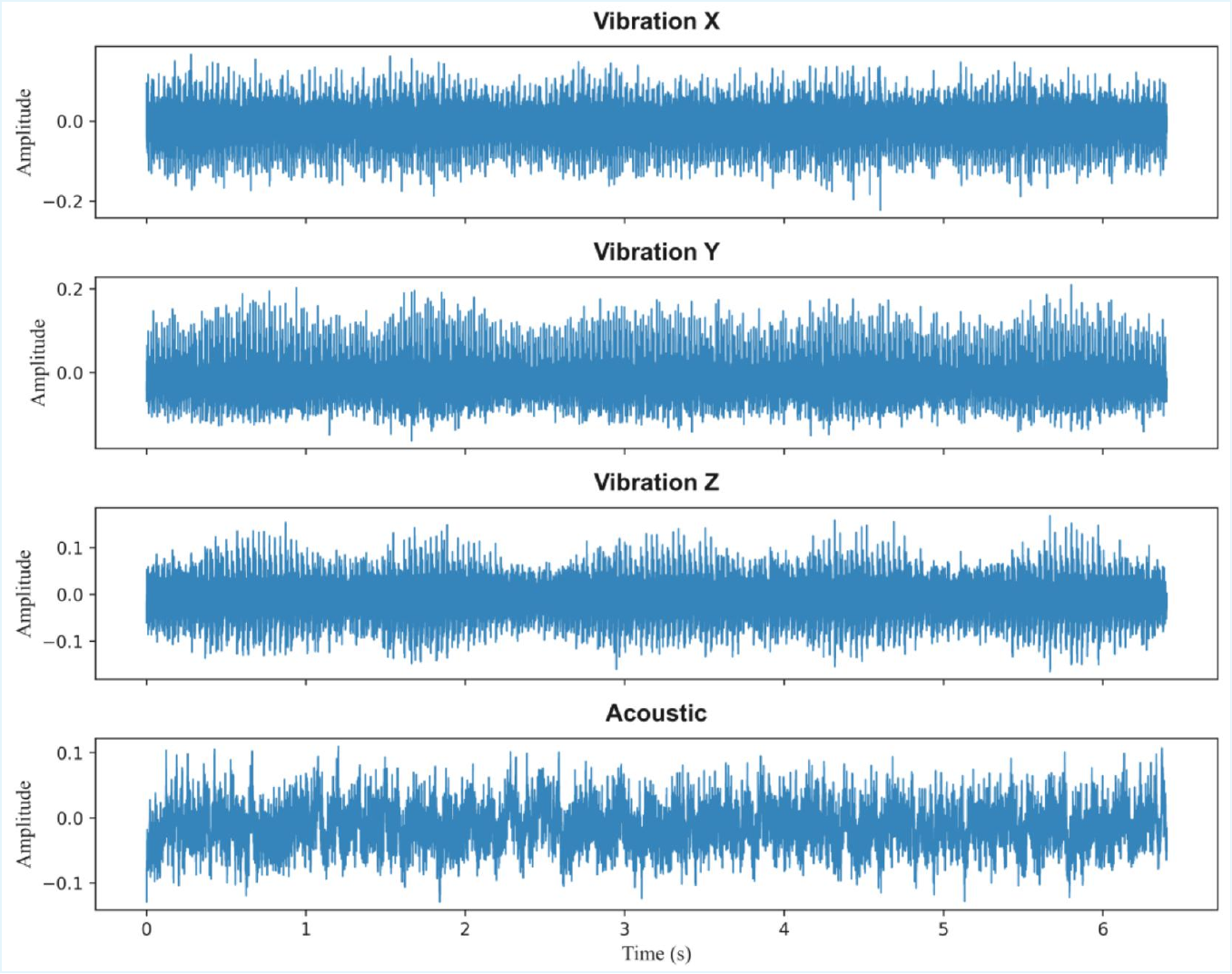}
  \vspace{-1em}
   \caption{Examples from fault diagnosis dataset. }
   \label{fig:fd}
\end{figure}

\noindent\textbf{Human-Animal-Cartoon (HAC)~\cite{dong2023SimMMDG}.} The HAC dataset consists of seven actions (“sleeping,” “watching TV,” “eating,” “drinking,” “swimming,” “running,” and “opening door”) performed by humans, animals, and cartoon characters, forming three distinct domains: Human (H), Animal (A), and Cartoon (C). The dataset contains a total of $3,381$ video clips, including $1,387$ human samples, $906$ animal samples, and $1,088$ cartoon samples. Each sample includes video, audio, and pre-computed optical flow modalities.

\noindent\textbf{EPIC-Kitchens~\cite{Munro_2020_CVPR}.} Following the experimental protocol of prior work~\cite{Munro_2020_CVPR}, we use a subset of EPIC-Kitchens containing eight actions (“put,” “take,” “open,” “close,” “wash,” “cut,” “mix,” and “pour”) recorded across three different kitchens, which define three domains: D1, D2, and D3. The dataset comprises $10,094$ video clips in total, with $1,978$ samples from D1, $3,245$ from D2, and $4,871$ from D3. Each sample includes video, audio, and pre-computed optical flow modalities.

\subsection{Mechanical Fault Diagnosis}

\noindent\textbf{HUST Motor~\cite{zhao2024domain}.} HUST Motor is a public motor fault diagnosis dataset that provides synchronized vibration and acoustic signals collected from a Spectra-Quest Mechanical Fault Simulator, distinguishing it from the predominantly vibration-only datasets commonly used in this field. The dataset covers six motor health states: healthy, bearing fault, bowed rotor, broken rotor bars, rotor misalignment, and voltage unbalance, with all faults artificially introduced to ensure controlled ground-truth labels. Each health condition is recorded under four steady-state rotational speeds (5, 10, 20, and 30 Hz), forming four distinct domains. Both vibration and acoustic signals are sampled at 25.6 kHz, with 163,840 samples collected for each configuration. The combination of complementary modalities, multiple operating conditions, and diverse fault categories makes HUST Motor a valuable benchmark for multimodal domain generalization in fault diagnosis.

\subsection{Sentiment Analysis}

\begin{figure}[t]
  \centering  \includegraphics[width=0.9\linewidth]{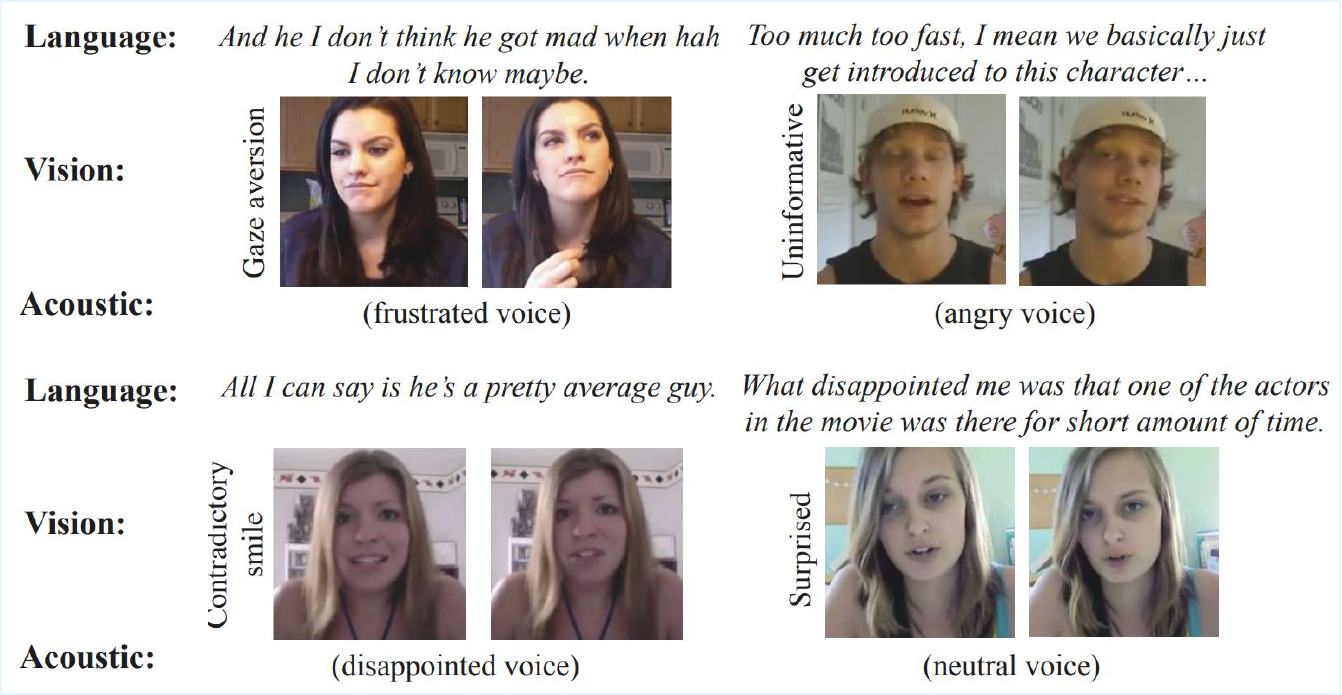}
  \vspace{-1em}
   \caption{Examples from sentiment analysis datasets. }
   \label{fig:sa}
\end{figure}

\noindent\textbf{CMU-MOSI~\cite{zadeh2016mosi}.} CMU-MOSI is a foundational dataset for English-language multimodal sentiment analysis. It contains 2,199 short opinion video clips collected from YouTube monologue reviews, with each utterance annotated for sentiment intensity on a continuous scale from -3 (highly negative) to +3 (highly positive). The dataset provides three temporally aligned modalities: text (transcribed speech), acoustic features (e.g., pitch and energy), and visual features (e.g., facial expressions and gestures). Despite its relatively modest scale, CMU-MOSI remains a widely used benchmark for sentiment regression and classification, and serves as a standard evaluation dataset for multimodal fusion methods. 

\noindent\textbf{CMU-MOSEI~\cite{zadeh2018multimodal}.} CMU-MOSEI extends CMU-MOSI and is one of the largest publicly available datasets for multimodal sentiment and emotion analysis. It includes over 23,500 sentence-level video utterances from more than 1,000 distinct YouTube speakers across diverse topics. Each utterance is annotated with both a sentiment intensity score in [-3, +3] and six Ekman-style emotion categories (happiness, sadness, anger, fear, disgust, and surprise) with corresponding intensity labels. Like CMU-MOSI, it provides temporally aligned text, acoustic, and visual modalities. Its large scale, speaker diversity, and comprehensive annotations make CMU-MOSEI a standard benchmark for multimodal fusion, transfer learning, and generalization research. Since CMU-MOSEI is a broader and larger-scale extension of CMU-MOSI, we only consider the MOSI $\rightarrow$ MOSEI generalization direction in our experiments.

\noindent\textbf{CH-SIMS~\cite{yu2020ch}.} CH-SIMS is a Chinese-language multimodal sentiment analysis dataset designed to address limitations of prior datasets that provide only unified multimodal sentiment labels. It consists of 2,281 refined video segments collected from real-world sources such as movies, TV series, and variety shows. In addition to an overall multimodal sentiment label, CH-SIMS provides independent sentiment annotations for each modality—text, audio, and visual—using a five-point scale ranging from negative to positive in [-1, +1]. This modality-specific labeling enables more detailed analysis of inter-modality consistency and disagreement, while also supporting unimodal, multimodal, and multi-task learning research. 

Since sentiment scales vary across datasets, we formulate sentiment classification as a binary task (negative vs. positive) and normalize regression targets to the range [-3, +3].

\section{Hyperparameter Spaces}
\label{sec:hyper}

We list all hyperparameters, their default values, and the corresponding search distributions used in our random hyperparameter sweeps in Table~\ref{tab:hyper}.

\begin{table}[t]
\centering
\caption{Hyperparameters, their default values and distributions for random search.}
\begin{tabular}{llll}
\toprule
\textbf{Condition} & \textbf{Parameter} & \textbf{Default value} & \textbf{Random distribution} \\
\midrule
RNA-Net & alpha\_RNA & 1.0 & {Uniform}(0, 3) \\
\midrule
\multirow{3}{*}{SimMMDG} & alpha\_trans & 0.1 & {Uniform}(0, 1) \\
 & explore\_loss\_coeff & 0.7 & {Uniform}(0, 1) \\
 & alpha\_contrast & 3.0 & {Uniform}(0, 5) \\

\midrule
\multirow{3}{*}{MOOSA} & entropy\_min\_weight & 0.001 & {Uniform}(0, 1) \\
 & jigsaw\_ratio & 1.0 & {Uniform}(0, 3) \\
 & mask\_ratio & 0.3 & {Uniform}(0, 1) \\
\midrule
\multirow{2}{*}{CMRF} & distill\_coef & 3.0 & {Uniform}(0, 5) \\
 & mix\_coef & 2.0 & {Uniform}(0, 5) \\
\midrule
\multirow{5}{*}{NEL} & alpha & 0.7 & {Uniform}(0, 1) \\
 & beta & 1/bsz & {Uniform}(0, 1) \\
 & temp\_s & 0.1 & {Uniform}(0, 1) \\
 & temp\_u & 0.25 & {Uniform}(0, 1) \\
 & k & 8 & {Choice}(\{4, 8, 12, 16\}) \\
\midrule
\multirow{5}{*}{JAT} & alpha\_rev & 0.1 & {Uniform}(0, 1) \\
 & alpha\_rev2 & 0.3 & {Uniform}(0, 1) \\
 & domain\_adv\_loss & 0.5 & {Uniform}(0, 1) \\
 & modal\_adv\_loss & 0.1 & {Uniform}(0, 1) \\
 & cls\_loss & 3.0 & {Uniform}(0, 5) \\
\midrule
\multirow{4}{*}{MBCD} & ema\_beta & 0.999 & {Uniform}(0.9, 1.0) \\
 & kl\_mm\_coeff & 1.0 & {Uniform}(0, 2) \\
 & kl\_um\_coeff & 1.0 & {Uniform}(0, 2) \\
 & modality\_drop\_base & 0 & {Uniform}(0, 1) \\
\midrule
\multirow{4}{*}{GMP} & alpha\_rev & 0.3 & {Uniform}(0, 1) \\
 & alpha\_k & 0.5 & {Uniform}(0, 1) \\
 & alpha\_p & 0.1 & {Uniform}(0, 1) \\
 & cls\_loss & 3.0 & {Uniform}(0, 5) \\
\bottomrule
\end{tabular}
\label{tab:hyper}
\end{table}

\section{Detailed Experimental Results}

We present detailed experimental results for single-source DG (Table \ref{tab:epic-ssdg} to Table \ref{tab:msa-ssdg}), as well as under corruption(Table \ref{tab:hac-dg-corruption-merged}) and missing-modality settings (Table \ref{tab:hac-dg-missing}).

\begin{table*}[t!]
\centering
\caption{Multimodal \textbf{single-source} DG  with different modalities on EPIC-Kitchens dataset.}
\vspace{0.4em}
\resizebox{\linewidth}{!}{
\begin{threeparttable}
\begin{tabular}{lccccccccccc}
\toprule
& \multicolumn{3}{c}{\textbf{Modality}} & \multicolumn{2}{c}{\textbf{Source: D1}}& \multicolumn{2}{c}{\textbf{Source: D2}}& \multicolumn{2}{c}{\textbf{Source: D3}}\\
\cmidrule(lr){2-4} \cmidrule(lr){5-6} \cmidrule(lr){7-8} \cmidrule(lr){9-10}
\textbf{Method} & Video & Audio & Flow & D1$\rightarrow$ D2 & D1$\rightarrow$ D3 & D2 $\rightarrow$ D1 & D2 $\rightarrow$ D3& D3$\rightarrow$ D1& D3$\rightarrow$ D2  & \textit{Mean}\\
\midrule

ERM & $\checkmark$& $\checkmark$&    & 51.07 & \secnd{54.72} & 43.45 & 55.44 & 46.67 & 56.13 & 51.25 \\
RNA-Net~\cite{Planamente_2022_WACV} & $\checkmark$& $\checkmark$&  & 52.53 & 51.85 & 51.03 & 56.26 & 53.79 & 55.60 & 53.51 \\
SimMMDG~\cite{dong2023SimMMDG}& $\checkmark$& $\checkmark$&  & 53.33 & 51.54 & \secnd{51.72} & 60.16 & \secnd{55.63} & \secnd{58.93} & \secnd{55.22} \\
MOOSA~\cite{dong2024moosa}& $\checkmark$& $\checkmark$&  & 53.60 & 51.23 & 47.82 & \secnd{61.91} & \best{56.55} & 58.80 & 54.98 \\
CMRF~\cite{fan2024cross}& $\checkmark$& $\checkmark$&  & \best{58.67} & 51.33 & 49.66 & \best{62.01} & 50.11 & 57.73 & 54.92 \\
NEL~\cite{zhang2025nonpolarized}& $\checkmark$& $\checkmark$&   & 54.66 & 54.07 & 47.81 & 59.13 & 48.50 & 57.51 & 53.61 \\
JAT~\cite{li2025towards}& $\checkmark$& $\checkmark$&   & 55.32 & 50.08 & 50.12 & 59.23 & 50.18 & 56.22 & 53.52 \\
MBCD~\cite{wang2025modality}& $\checkmark$& $\checkmark$&   & \secnd{56.22} & \best{55.30} & \best{53.41} & 61.17 & 53.64 & \best{62.26} & \best{57.00} \\
GMP~\cite{li2026balancing} & $\checkmark$& $\checkmark$& & 53.17 & 49.82 & 48.97 & 59.65 & 49.81 & 57.33 & 53.12 \\
\textit{Oracle} & $\checkmark$& $\checkmark$&  & \textit{76.13} & \textit{76.80} & \textit{60.23} & \textit{76.80} & \textit{60.23} & \textit{76.13} & \textit{71.05} \\

\midrule

ERM & $\checkmark$& &$\checkmark$ & \secnd{58.93} & \best{55.24} & 49.43 & 56.98 & 55.40 & 64.40 & 56.73 \\
RNA-Net~\cite{Planamente_2022_WACV} & $\checkmark$& & $\checkmark$ & 56.40 & \secnd{54.93} & 53.56 & 58.01 & \secnd{56.78} & 62.27 & 56.99 \\
SimMMDG~\cite{dong2023SimMMDG} & $\checkmark$& & $\checkmark$ & \best{59.07} & 51.13 & \best{56.55} & 59.14 & \best{57.93} & 64.27 & \secnd{58.01} \\
MOOSA~\cite{dong2024moosa}& $\checkmark$& &  $\checkmark$ & 57.07 & 50.51 & 54.25 & \secnd{62.22} & 54.94 & \secnd{66.00} & 57.50 \\
CMRF~\cite{fan2024cross}& $\checkmark$& & $\checkmark$  & 58.27 & 49.79 & 52.64 & 60.27 & 56.09 & 64.00 & 56.84 \\
NEL~\cite{zhang2025nonpolarized}& $\checkmark$& &   $\checkmark$ & 54.75 & 47.43 & 52.79 & 60.95 & 54.40 & 63.11 & 55.57 \\
JAT~\cite{li2025towards}& $\checkmark$& &   $\checkmark$ & 55.32 & 48.18 & 53.16 & 59.02 & 55.78 & 63.55 & 55.84 \\
MBCD~\cite{wang2025modality}& $\checkmark$& & $\checkmark$  & 56.31 & 53.18 & \secnd{55.55} & \best{62.28} & 56.55 & \best{67.58} & \best{58.57} \\
GMP~\cite{li2026balancing} & $\checkmark$& & $\checkmark$ & 54.83 & 50.67 & 51.67 & 59.19 & 55.82 & 64.19 & 56.06 \\
\textit{Oracle} & $\checkmark$& & $\checkmark$ & \textit{80.00} & \textit{81.21} & \textit{65.52} & \textit{81.21} & \textit{65.52} & \textit{80.00} & \textit{75.58} \\

\midrule

ERM & & $\checkmark$&$\checkmark$ & 47.20 & 49.38 & 42.53 & 52.57 & 47.13 & 57.33 & 49.36 \\
RNA-Net~\cite{Planamente_2022_WACV} & & $\checkmark$&$\checkmark$ & 50.93 & \secnd{54.00} & 42.07 & 54.72 & 48.51 & 57.87 & 51.35 \\
SimMMDG~\cite{dong2023SimMMDG} & & $\checkmark$&$\checkmark$ & \best{53.47} & 51.33 & \secnd{47.13} & \secnd{56.06} & 52.64 & \secnd{63.33} & 53.99 \\
MOOSA~\cite{dong2024moosa}& & $\checkmark$&  $\checkmark$ & \secnd{53.07} & \best{54.11} & 45.75 & 55.75 & \best{54.48} & 63.07 & \secnd{54.37} \\
CMRF~\cite{fan2024cross}&  & $\checkmark$&   $\checkmark$ & 50.93 & 53.59 & 43.22 & 52.87 & 49.89 & 62.40 & 52.15 \\
NEL~\cite{zhang2025nonpolarized}&  & $\checkmark$& $\checkmark$  & 49.91 & 50.34 & 44.13 & \best{57.46} & 50.19 & 60.08 & 52.02 \\
JAT~\cite{li2025towards}&  & $\checkmark$& $\checkmark$  & 50.09 & 52.11 & 43.29 & 54.66 & 50.17 & 59.89 & 51.70 \\
MBCD~\cite{wang2025modality}& & $\checkmark$&   $\checkmark$ & 52.62 & 53.73 & \best{54.35} & 52.94 & \secnd{54.25} & \best{66.02} & \best{55.65} \\
GMP~\cite{li2026balancing} & & $\checkmark$& $\checkmark$ & 49.83 & 50.25 & 44.32 & 53.88 & 49.64 & 58.71 & 51.10 \\
\textit{Oracle} & & $\checkmark$& $\checkmark$ & \textit{74.13} & \textit{73.61} & \textit{59.77} & \textit{73.61} & \textit{59.77} & \textit{74.13} & \textit{69.17} \\

\midrule

ERM & $\checkmark$& $\checkmark$& $\checkmark$ & 55.47 & 52.87 & 52.64 & 58.52 & 55.86 & 63.60 & 56.49 \\
RNA-Net~\cite{Planamente_2022_WACV} & $\checkmark$& $\checkmark$& $\checkmark$ & 59.07 & 56.06 & \secnd{53.10} & 60.16 & 52.64 & 64.80 & 57.64 \\
SimMMDG~\cite{dong2023SimMMDG} & $\checkmark$& $\checkmark$& $\checkmark$ & 58.27 & 53.49 & 51.49 & \secnd{63.35} & \secnd{58.16} & \secnd{70.93} & 59.28 \\
MOOSA~\cite{dong2024moosa}& $\checkmark$& $\checkmark$& $\checkmark$  & \best{60.27} & \best{57.39} & 50.57 & 62.53 & \best{61.15} & 66.27 & \secnd{59.70} \\
CMRF~\cite{fan2024cross}& $\checkmark$& $\checkmark$&$\checkmark$   & 59.47 & \secnd{56.37} & 51.72 & 61.29 & 57.01 & 66.40 & 58.71 \\
NEL~\cite{zhang2025nonpolarized}& $\checkmark$& $\checkmark$&  $\checkmark$ & 58.40 & 54.07 & 49.19 & 62.25 & 55.25 & 66.04 & 57.53 \\
JAT~\cite{li2025towards}& $\checkmark$& $\checkmark$& $\checkmark$  & 58.61 & 54.35 & 50.16 & 61.33 & 56.38 & 63.24 & 57.34 \\
MBCD~\cite{wang2025modality}& $\checkmark$& $\checkmark$& $\checkmark$  & \secnd{60.04} & 55.91 & \best{55.78} & \best{64.81} & 56.78 & \best{72.00} & \best{60.89} \\
GMP~\cite{li2026balancing} & $\checkmark$& $\checkmark$& $\checkmark$ & 57.38 & 53.59 & 50.59 & 61.55 & 54.82 & 65.79 & 57.29 \\
\textit{Oracle} & $\checkmark$& $\checkmark$& $\checkmark$ & \textit{79.47} & \textit{78.64} & \textit{65.52} & \textit{78.64} & \textit{65.52} & \textit{79.47} & \textit{74.54} \\
\bottomrule
\end{tabular}
\end{threeparttable}
}
\label{tab:epic-ssdg}
\end{table*}
\begin{table*}[t!]
\centering
\caption{Multimodal \textbf{single-source} DG  with different modalities on HAC dataset.}
\vspace{0.4em}
\resizebox{\linewidth}{!}{
\begin{threeparttable}
\begin{tabular}{lccccccccccc}
\toprule
& \multicolumn{3}{c}{\textbf{Modality}} & \multicolumn{2}{c}{\textbf{Source: H}}& \multicolumn{2}{c}{\textbf{Source: A}}& \multicolumn{2}{c}{\textbf{Source: C}}\\
\cmidrule(lr){2-4} \cmidrule(lr){5-6} \cmidrule(lr){7-8} \cmidrule(lr){9-10}
\textbf{Method} & Video & Audio & Flow & H$\rightarrow$ A & H$\rightarrow$ C & A $\rightarrow$ H & A $\rightarrow$ C& C$\rightarrow$ H& C$\rightarrow$ A  & \textit{Mean}\\

\midrule

ERM & $\checkmark$& $\checkmark$&    & 66.67 & 49.36 & 65.83 & 50.00 & 64.67 & \secnd{72.74} & 61.54 \\
RNA-Net~\cite{Planamente_2022_WACV} & $\checkmark$& $\checkmark$&  & 65.89 & \best{52.11} & 67.84 & \secnd{53.13} & 60.27 & 71.30 & 61.76 \\
SimMMDG~\cite{dong2023SimMMDG}& $\checkmark$& $\checkmark$&   & 68.21 & 45.86 & \best{75.34} & 50.64 & 69.00 & \best{73.18} & \secnd{63.71} \\
MOOSA~\cite{dong2024moosa}& $\checkmark$& $\checkmark$&   & 67.99 & 43.38 & 72.39 & 49.45 & \secnd{70.87} & 72.08 & 62.69 \\
CMRF~\cite{fan2024cross}& $\checkmark$& $\checkmark$&   & 66.78 & 45.59 & 73.54 & \best{54.96} & \best{74.55} & 71.52 & \best{64.49} \\
NEL~\cite{zhang2025nonpolarized}& $\checkmark$& $\checkmark$&  & \secnd{68.57} & 46.32 & \secnd{74.90} & 45.52 & 69.50 & 69.31 & 62.35 \\
JAT~\cite{li2025towards}& $\checkmark$& $\checkmark$&   & 66.84 & 44.15 & 70.31 & 45.28 & 65.51 & 70.82 & 60.49 \\
MBCD~\cite{wang2025modality}& $\checkmark$& $\checkmark$& & \best{68.69} & 42.93 & 71.52 & 43.35 & 65.97 & 69.57 & 60.34 \\
GMP~\cite{li2026balancing} & $\checkmark$& $\checkmark$&  & 67.29 & 48.71 & 72.43 & 44.48 & 64.65 & 69.13 & 61.12 \\
\textit{Oracle} & $\checkmark$& $\checkmark$&  & \textit{97.16} & \textit{88.53} & \textit{92.75} & \textit{88.53} & \textit{92.75} & \textit{97.16} & \textit{92.81} \\

\midrule

ERM & $\checkmark$& &$\checkmark$ & 65.78 & 45.31 & 75.78 & 48.35 & 69.79 & 64.13 & 61.52 \\
RNA-Net~\cite{Planamente_2022_WACV} & $\checkmark$& & $\checkmark$ & 64.90 & \secnd{45.40} & 72.24 & 50.09 & 59.63 & 65.01 & 59.54 \\
SimMMDG~\cite{dong2023SimMMDG} & $\checkmark$& & $\checkmark$ & \best{68.87} & 43.84 & 74.33 & \secnd{53.13} & \secnd{71.23} & 65.12 & 62.75 \\
MOOSA~\cite{dong2024moosa}& $\checkmark$& &  $\checkmark$ & \secnd{67.99} & 45.31 & 76.42 & \best{54.04} & 70.37 & \best{68.10} & \best{63.71} \\
CMRF~\cite{fan2024cross}& $\checkmark$& & $\checkmark$  & 66.78 & \best{47.61} & \secnd{77.22} & 51.93 & 69.72 & \secnd{66.56} & \secnd{63.30} \\
NEL~\cite{zhang2025nonpolarized}& $\checkmark$& &   $\checkmark$ & 65.85 & 41.14 & 73.61 & 36.58 & 69.14 & 65.12 & 58.57 \\
JAT~\cite{li2025towards}& $\checkmark$& &   $\checkmark$ & 66.37 & 40.91 & 71.37 & 41.66 & 62.86 & 59.30 & 57.08 \\
MBCD~\cite{wang2025modality}& $\checkmark$& & $\checkmark$  & 67.88 & 40.50 & \best{79.23} & 48.77 & \best{71.32} & 62.62 & 61.72 \\
GMP~\cite{li2026balancing} & $\checkmark$& & $\checkmark$ & 66.74 & 41.94 & 69.21 & 41.91 & 63.54 & 60.49 & 57.30 \\
\textit{Oracle} & $\checkmark$& & $\checkmark$ & \textit{96.59} & \textit{85.78} & \textit{93.48} & \textit{85.78} & \textit{93.48} & \textit{96.59} & \textit{91.95} \\

\midrule

ERM & & $\checkmark$&$\checkmark$ & 57.73 & \best{40.35} & 52.49 & 39.61 & 38.57 & 49.34 & 46.35 \\
RNA-Net~\cite{Planamente_2022_WACV} & & $\checkmark$&$\checkmark$ & 54.86 & 38.05 & 50.32 & \secnd{44.21} & 41.17 & 49.67 & 46.38 \\
SimMMDG~\cite{dong2023SimMMDG} & & $\checkmark$&$\checkmark$ & \best{61.81} & \secnd{40.07} & 56.02 & 41.73 & 41.89 & 50.00 & 48.59 \\
MOOSA~\cite{dong2024moosa}& & $\checkmark$&  $\checkmark$ & 59.38 & 39.98 & 58.54 & \best{44.49} & 40.88 & \best{52.87} & \secnd{49.36} \\
CMRF~\cite{fan2024cross}&  & $\checkmark$&   $\checkmark$ & 58.06 & 39.98 & \secnd{58.69} & 43.38 & \secnd{43.76} & 45.92 & 48.30 \\
NEL~\cite{zhang2025nonpolarized}&  & $\checkmark$& $\checkmark$  & 57.43 & 34.98 & 56.16 & 36.76 & 38.50 & 44.44 & 44.71 \\
JAT~\cite{li2025towards}&  & $\checkmark$& $\checkmark$  & 56.80 & 36.80 & 50.73 & 37.31 & 38.55 & 46.06 & 44.38 \\
MBCD~\cite{wang2025modality}& & $\checkmark$&   $\checkmark$ & \secnd{60.15} & 36.33 & \best{58.80} & 41.69 & \best{50.03} & \secnd{51.28} & \best{49.71} \\
GMP~\cite{li2026balancing} & & $\checkmark$& $\checkmark$ & 57.26 & 37.88 & 54.37 & 38.61 & 40.15 & 48.17 & 46.07 \\
\textit{Oracle} & & $\checkmark$& $\checkmark$ & \textit{90.91} & \textit{68.35} & \textit{81.52} & \textit{68.35} & \textit{81.52} & \textit{90.91} & \textit{80.26} \\

\midrule

ERM & $\checkmark$& $\checkmark$& $\checkmark$ & \secnd{68.10} & 44.67 & 70.44 & 50.83 & 63.30 & 68.43 & 60.96 \\
RNA-Net~\cite{Planamente_2022_WACV} & $\checkmark$& $\checkmark$& $\checkmark$ & 64.35 & 46.23 & 67.27 & 48.99 & 61.93 & 65.45 & 59.04 \\
SimMMDG~\cite{dong2023SimMMDG} & $\checkmark$& $\checkmark$& $\checkmark$ & 66.45 & 45.13 & \secnd{73.90} & 52.30 & \best{70.44} & \best{73.40} & \best{63.60} \\
MOOSA~\cite{dong2024moosa}& $\checkmark$& $\checkmark$& $\checkmark$  & 66.11 & 47.98 & 72.03 & \secnd{52.67} & 66.33 & \secnd{72.74} & 62.98 \\
CMRF~\cite{fan2024cross}& $\checkmark$& $\checkmark$&$\checkmark$  & 67.99 & \best{49.54} & 69.36 & \best{55.70} & 65.10 & 65.12 & 62.13 \\
NEL~\cite{zhang2025nonpolarized}& $\checkmark$& $\checkmark$&  $\checkmark$ & 65.92 & 43.72 & 70.22 & 47.70 & 65.54 & 66.15 & 59.88 \\
JAT~\cite{li2025towards}& $\checkmark$& $\checkmark$& $\checkmark$  & 65.27 & 45.58 & 70.75 & 45.77 & 61.06 & 64.09 & 58.75 \\
MBCD~\cite{wang2025modality}& $\checkmark$& $\checkmark$& $\checkmark$  & \best{70.42} & 44.45 & \best{77.21} & 49.63 & \secnd{70.17} & 69.28 & \secnd{63.53} \\
GMP~\cite{li2026balancing} & $\checkmark$& $\checkmark$& $\checkmark$ & 63.42 & \secnd{48.26} & 71.18 & 47.30 & 63.97 & 62.51 & 59.44 \\
\textit{Oracle} & $\checkmark$& $\checkmark$& $\checkmark$ & \textit{96.02} & \textit{86.24} & \textit{92.75} & \textit{86.24} & \textit{92.75} & \textit{96.02} & \textit{91.67} \\
\bottomrule
\end{tabular}

\end{threeparttable}
}
\label{tab:hac-ssdg-appen}
\end{table*}

\begin{table*}[t!]
\centering
\caption{Multimodal \textbf{single-source} DG on HUST dataset with vibration and acoustic modalities.}
\vspace{0.4em}
\resizebox{\linewidth}{!}{
\begin{threeparttable}
\begin{tabular}{lcccccccccccccc}
\toprule
& \multicolumn{3}{c}{\textbf{Source: D1}} & \multicolumn{3}{c}{\textbf{Source: D2}} & \multicolumn{3}{c}{\textbf{Source: D3}} & \multicolumn{3}{c}{\textbf{Source: D4}} \\
\cmidrule(lr){2-4} \cmidrule(lr){5-7} \cmidrule(lr){8-10} \cmidrule(lr){11-13}
\textbf{Method} \hspace{2mm} Target: & D2 & D3 & D4 & D1 & D3 & D4 & D1 & D2 & D4 & D1 & D2 & D3 & \textit{Mean} \\
\midrule
ERM & 45.08 & 12.00 & 4.92 & 50.00 & 58.92 & 50.67 & 21.42 & 58.25 & 80.17 & 18.00 & 37.83 & 74.25 & 42.63 \\
RNA-Net~\cite{Planamente_2022_WACV} & 56.08 & 12.83 & 1.75 & 51.58 & 52.08 & 48.75 & \best{27.00} & 62.92 & 80.92 & \best{21.08} & 40.83 & 74.33 & 44.18 \\
SimMMDG~\cite{dong2023SimMMDG} & 51.25 & 16.00 & 14.67 & 44.75 & \best{68.83} & \best{63.17} & 24.50 & \best{71.75} & 81.83 & 17.00 & \best{48.00} & 76.67 & \best{48.20} \\
MOOSA~\cite{dong2024moosa} & 49.50 & 18.58 & 11.50 & \best{55.92} & \secnd{66.83} & 58.00 & \secnd{26.67} & 63.17 & \secnd{82.83} & 14.67 & \secnd{46.92} & 75.25 & \secnd{47.49} \\
CMRF~\cite{fan2024cross} & 49.50 & 17.58 & 16.67 & 51.42 & 60.42 & 46.33 & 21.17 & 51.25 & 76.50 & 17.25 & 42.25 & \best{78.83} & 44.10 \\
NEL~\cite{zhang2025nonpolarized} & 57.39 & \secnd{20.92} & \secnd{17.86} & \secnd{53.39} & 56.94 & 47.17 & 24.86 & 52.00 & 75.89 & \secnd{18.39} & 32.36 & 72.28 & 44.12 \\
JAT~\cite{li2025towards} & 58.20 & 16.39 & 10.95 & 44.19 & 58.78 & 57.70 & 22.06 & 58.42 & 80.67 & 18.17 & 42.64 & 76.36 & 45.38 \\
MBCD~\cite{wang2025modality} & \best{63.33} & \best{23.19} & \best{18.14} & 38.25 & 66.81 & 53.94 & 17.69 & \secnd{66.58} & 66.72 & 15.25 & 43.55 & 70.11 & 45.30 \\
GMP~\cite{li2026balancing} & \secnd{58.67} & 17.36 & 14.19 & 50.11 & 63.78 & \secnd{59.33} & 17.00 & 58.03 & \best{82.92} & 17.97 & 39.05 & \secnd{77.61} & 46.34 \\
\textit{Oracle} & \textit{99.83} & \textit{100.00} & \textit{99.83} & \textit{99.83} & \textit{100.00} & \textit{99.83} & \textit{99.83} & \textit{99.83} & \textit{99.83} & \textit{99.83} & \textit{99.83} & \textit{100.00} & \textit{99.87} \\
\bottomrule
\end{tabular}
\end{threeparttable}
}
\label{tab:hust-ssdg}
\end{table*}

\begin{table*}[t!]
\centering
\caption{Multimodal \textbf{single-source} DG on MOSI, MOSEI, and SIMS datasets for sentiment analysis with video, audio, and text modalities.}
\vspace{0.4em}
\setlength{\tabcolsep}{2.5pt}
\renewcommand{\arraystretch}{1.1}
\resizebox{\linewidth}{!}{
\begin{threeparttable}
\begin{tabular}{l ccc ccc ccc ccc ccc ccc}
\toprule
\textbf{Method} & \multicolumn{3}{c}{MOSEI $\rightarrow$ SIMS} & \multicolumn{3}{c}{MOSI $\rightarrow$ SIMS} & \multicolumn{3}{c}{MOSI $\rightarrow$ MOSEI} & \multicolumn{3}{c}{SIMS $\rightarrow$ MOSI} & \multicolumn{3}{c}{SIMS $\rightarrow$ MOSEI} & \multicolumn{3}{c}{\textit{Mean}} \\
\cmidrule(lr){2-4} \cmidrule(lr){5-7} \cmidrule(lr){8-10} \cmidrule(lr){11-13} \cmidrule(lr){14-16} \cmidrule(lr){17-19}
& MAE$\downarrow$ & F1$\uparrow$ & ACC2$\uparrow$ & MAE$\downarrow$ & F1$\uparrow$ & ACC2$\uparrow$ & MAE$\downarrow$ & F1$\uparrow$ & ACC2$\uparrow$ & MAE$\downarrow$ & F1$\uparrow$ & ACC2$\uparrow$ & MAE$\downarrow$ & F1$\uparrow$ & ACC2$\uparrow$ & MAE$\downarrow$ & F1$\uparrow$ & ACC2$\uparrow$ \\
\midrule
ERM & \best{1.79} & 67.92 & 63.68 & \best{1.81} & 67.74 & 62.80 & 0.99 & 67.55 & 66.96 & \secnd{1.46} & 74.61 & \secnd{60.03} & 1.60 & 66.39 & 50.55 & 1.53 & 68.84 & 60.80 \\
RNA-Net~\cite{Planamente_2022_WACV} & 1.81 & 72.41 & 66.30 & 1.85 & 66.69 & 61.71 & 0.93 & \secnd{68.12} & \secnd{68.10} & 1.49 & 73.49 & 59.29 & 1.43 & 65.86 & 50.51 & 1.50 & 69.31 & 61.18 \\
SimMMDG~\cite{dong2023SimMMDG} & \secnd{1.80} & \best{75.94} & \best{66.96} & 1.89 & \best{75.55} & \best{68.05} & 0.98 & \best{68.86} & \best{68.60} & \secnd{1.46} & \best{74.91} & 59.88 & 1.42 & 66.65 & \best{51.02} & 1.51 & \best{72.38} & \best{62.90} \\
MOOSA~\cite{dong2024moosa} & 1.83 & \secnd{75.93} & \secnd{66.52} & \best{1.81} & 63.30 & 60.61 & \secnd{0.89} & 67.33 & 67.22 & 1.59 & 74.06 & \best{60.18} & 1.39 & \secnd{66.69} & 50.59 & 1.50 & 69.46 & 61.02 \\
CMRF~\cite{fan2024cross} & 1.84 & 74.07 & 65.65 & 1.86 & 70.91 & 63.02 & 0.94 & 66.96 & 66.96 & \best{1.38} & \secnd{74.87} & \secnd{60.03} & \best{0.94} & \best{67.11} & \secnd{50.92} & \best{1.39} & \secnd{70.78} & \secnd{61.32} \\
NEL~\cite{zhang2025nonpolarized} & \best{1.79} & 64.64 & 63.09 & 1.85 & 64.78 & 61.39 & 1.03 & 67.17 & 66.87 & 1.47 & 73.87 & 59.53 & 1.42 & 61.03 & 50.51 & 1.51 & 66.30 & 60.28 \\
JAT~\cite{li2025towards} & 1.88 & 70.64 & 64.99 & 1.89 & 69.11 & 62.80 & 1.00 & 67.64 & 67.43 & 1.51 & 74.83 & 59.83 & 1.39 & 56.78 & 46.95 & 1.53 & 67.80 & 60.40 \\
MBCD~\cite{wang2025modality} & 1.98 & 51.28 & 49.22 & \secnd{1.82} & \secnd{73.74} & \secnd{66.07} & \best{0.85} & 65.74 & 64.81 & 1.53 & 70.90 & 58.06 & \secnd{1.12} & 53.03 & 48.66 & \secnd{1.46} & 62.94 & 57.36 \\
GMP~\cite{li2026balancing} & 1.84 & 65.18 & 61.77 & \secnd{1.82} & 71.15 & 64.47 & 1.06 & 66.08 & 65.30 & 1.56 & 69.99 & 57.86 & 1.23 & 57.01 & 50.84 & 1.50 & 65.88 & 60.05 \\
\textit{Oracle} & \textit{1.32} & \textit{76.80} & \textit{76.80} & \textit{1.32} & \textit{76.80} & \textit{76.80} & \textit{0.58} & \textit{73.89} & \textit{73.63} & \textit{0.97} & \textit{78.37} & \textit{78.47} & \textit{0.58} & \textit{73.89} & \textit{73.63} & \textit{0.95} & \textit{75.95} & \textit{75.87} \\
\bottomrule
\end{tabular}
\end{threeparttable}
}
\label{tab:msa-ssdg}
\end{table*}

\begin{table*}[t!]
\centering
\caption{Multimodal multi-source DG with \textbf{corruptions} on HAC dataset. Subscripts show the change relative to the clean Video+Audio setting.}
\vspace{0.4em}
\resizebox{0.75\linewidth}{!}{
\begin{threeparttable}
\begin{tabular}{llcccccccccccc}
\toprule
 & & \multicolumn{4}{c}{\textbf{HAC dataset}}\\
\cmidrule(lr){3-6}
\textbf{Corruption} & \textbf{Method}  &  A, C $\rightarrow$ H & H, C $\rightarrow$ A & H, A $\rightarrow$ C  & \textit{Mean}\\
\midrule
\multirow{9}{*}{\shortstack{Wind\\ on audio}}
 & ERM     & 77.43$_{\textcolor{blue}{+1.52}}$ & 74.39$_{\textcolor{red}{-3.09}}$ & 52.67$_{\textcolor{red}{-0.73}}$ & 68.16$_{\textcolor{red}{-0.77}}$ \\
 & RNA-Net~\cite{Planamente_2022_WACV} & 75.78$_{\textcolor{blue}{+0.58}}$ & 73.51$_{\textcolor{red}{-3.97}}$ & 49.36$_{\textcolor{red}{-4.22}}$ & 66.22$_{\textcolor{red}{-2.53}}$ \\
 & SimMMDG~\cite{dong2023SimMMDG}& \best{79.96}$_{\textcolor{blue}{+1.37}}$ & \best{76.49}$_{\textcolor{red}{-1.55}}$ & \best{56.25}$_{\textcolor{blue}{+0.46}}$ & \best{70.90}$_{\textcolor{blue}{+0.09}}$ \\
 & MOOSA~\cite{dong2024moosa}& 77.65$_{\textcolor{red}{-1.73}}$ & 75.71$_{\textcolor{red}{-2.99}}$ & \secnd{54.78}$_{\textcolor{blue}{+0.00}}$ & \secnd{69.38}$_{\textcolor{red}{-1.57}}$ \\
 & CMRF~\cite{fan2024cross}& 76.06$_{\textcolor{red}{-1.88}}$ & 71.63$_{\textcolor{red}{-6.63}}$ & 47.70$_{\textcolor{red}{-4.14}}$ & 65.13$_{\textcolor{red}{-4.22}}$ \\
 & NEL~\cite{zhang2025nonpolarized}& 75.83$_{\textcolor{red}{-0.50}}$ & 73.96$_{\textcolor{red}{-2.46}}$ & 46.20$_{\textcolor{red}{-4.87}}$ & 65.33$_{\textcolor{red}{-2.61}}$ \\
 & JAT~\cite{li2025towards}& 76.42$_{\textcolor{red}{-1.74}}$ & 74.27$_{\textcolor{red}{-3.72}}$ & 49.82$_{\textcolor{red}{-3.29}}$ & 66.84$_{\textcolor{red}{-2.91}}$ \\
 & MBCD~\cite{wang2025modality}& \secnd{77.93}$_{\textcolor{red}{-0.19}}$ & \secnd{75.99}$_{\textcolor{red}{-2.92}}$ & 51.62$_{\textcolor{red}{-1.87}}$ & 68.51$_{\textcolor{red}{-1.66}}$ \\
 & GMP~\cite{li2026balancing} & 76.48$_{\textcolor{red}{-0.88}}$ & 74.92$_{\textcolor{red}{-1.55}}$ & 48.69$_{\textcolor{red}{-3.64}}$ & 66.70$_{\textcolor{red}{-2.02}}$ \\
\midrule
\multirow{9}{*}{\shortstack{Defocus\\ on video}}
 & ERM     & 65.54$_{\textcolor{red}{-10.37}}$ & 73.07$_{\textcolor{red}{-4.41}}$ & 41.82$_{\textcolor{red}{-11.58}}$ & 60.14$_{\textcolor{red}{-8.79}}$ \\
 & RNA-Net~\cite{Planamente_2022_WACV} & 61.93$_{\textcolor{red}{-13.27}}$ & \best{73.84}$_{\textcolor{red}{-3.64}}$ & 44.67$_{\textcolor{red}{-8.91}}$ & 60.15$_{\textcolor{red}{-8.60}}$ \\
 & SimMMDG~\cite{dong2023SimMMDG}& 61.50$_{\textcolor{red}{-17.09}}$ & 68.54$_{\textcolor{red}{-9.50}}$ & 43.93$_{\textcolor{red}{-11.86}}$ & 57.99$_{\textcolor{red}{-12.82}}$ \\
 & MOOSA~\cite{dong2024moosa}& \best{68.13}$_{\textcolor{red}{-11.25}}$ & 70.09$_{\textcolor{red}{-8.61}}$ & \best{48.44}$_{\textcolor{red}{-6.34}}$ & \best{62.22}$_{\textcolor{red}{-8.73}}$ \\
 & CMRF~\cite{fan2024cross}& 62.29$_{\textcolor{red}{-15.65}}$ & 70.42$_{\textcolor{red}{-7.84}}$ & 45.59$_{\textcolor{red}{-6.25}}$ & 59.43$_{\textcolor{red}{-9.92}}$ \\
 & NEL~\cite{zhang2025nonpolarized}& 60.74$_{\textcolor{red}{-15.59}}$ & 69.21$_{\textcolor{red}{-7.21}}$ & 40.29$_{\textcolor{red}{-10.78}}$ & 56.75$_{\textcolor{red}{-11.19}}$ \\
 & JAT~\cite{li2025towards}& 61.35$_{\textcolor{red}{-16.81}}$ & 68.66$_{\textcolor{red}{-9.33}}$ & 42.75$_{\textcolor{red}{-10.36}}$ & 57.59$_{\textcolor{red}{-12.16}}$ \\
 & MBCD~\cite{wang2025modality}& \secnd{65.88}$_{\textcolor{red}{-12.24}}$ & 72.65$_{\textcolor{red}{-6.26}}$ & \secnd{47.36}$_{\textcolor{red}{-6.13}}$ & \secnd{61.96}$_{\textcolor{red}{-8.21}}$ \\
 & GMP~\cite{li2026balancing} & 64.73$_{\textcolor{red}{-12.63}}$ & 70.95$_{\textcolor{red}{-5.52}}$ & 46.57$_{\textcolor{red}{-5.76}}$ & 60.75$_{\textcolor{red}{-7.97}}$ \\
\bottomrule
\end{tabular}
\end{threeparttable}
}
\label{tab:hac-dg-corruption-merged}
\end{table*}
\begin{table*}[t!]
\centering
\caption{Multimodal multi-source DG with \textbf{missing modalities} on HAC dataset. Subscripts show the change relative to the full Video+Audio setting.}
\vspace{0.4em}
\resizebox{0.8\linewidth}{!}{
\begin{threeparttable}
\begin{tabular}{lcccccccccccc}
\toprule
& \multicolumn{2}{c}{\textbf{Modality}} & \multicolumn{4}{c}{\textbf{HAC dataset}}\\
\cmidrule(lr){2-3} \cmidrule(lr){4-7} 
\textbf{Method} & Video & Audio &  A, C $\rightarrow$ H & H, C $\rightarrow$ A & H, A $\rightarrow$ C  & \textit{Mean}\\
\midrule
ERM & $\checkmark$& \ding{55} & 78.88$_{\textcolor{blue}{+2.97}}$ & 74.72$_{\textcolor{red}{-2.76}}$ & 47.70$_{\textcolor{red}{-5.70}}$ & 67.10$_{\textcolor{red}{-1.83}}$ \\
RNA-Net~\cite{Planamente_2022_WACV} & $\checkmark$& \ding{55} & 77.79$_{\textcolor{blue}{+2.59}}$ & 75.28$_{\textcolor{red}{-2.20}}$ & 49.72$_{\textcolor{red}{-3.86}}$ & 67.60$_{\textcolor{red}{-1.15}}$ \\
SimMMDG~\cite{dong2023SimMMDG}& $\checkmark$& \ding{55} & \best{80.82}$_{\textcolor{blue}{+2.23}}$ & \best{77.04}$_{\textcolor{red}{-1.00}}$ & \best{53.58}$_{\textcolor{red}{-2.21}}$ & \best{70.48}$_{\textcolor{red}{-0.33}}$ \\
MOOSA~\cite{dong2024moosa}& $\checkmark$& \ding{55} & 78.73$_{\textcolor{red}{-0.65}}$ & 76.49$_{\textcolor{red}{-2.21}}$ & \secnd{53.49}$_{\textcolor{red}{-1.29}}$ & 69.57$_{\textcolor{red}{-1.38}}$ \\
CMRF~\cite{fan2024cross}& $\checkmark$& \ding{55} & 79.81$_{\textcolor{blue}{+1.87}}$ & 71.41$_{\textcolor{red}{-6.85}}$ & 47.24$_{\textcolor{red}{-4.60}}$ & 66.15$_{\textcolor{red}{-3.20}}$ \\
NEL~\cite{zhang2025nonpolarized}& $\checkmark$& \ding{55} & 79.78$_{\textcolor{blue}{+3.45}}$ & 73.14$_{\textcolor{red}{-3.28}}$ & 47.09$_{\textcolor{red}{-3.98}}$ & 66.67$_{\textcolor{red}{-1.27}}$ \\
JAT~\cite{li2025towards}& $\checkmark$& \ding{55} & 79.29$_{\textcolor{blue}{+1.13}}$ & 74.32$_{\textcolor{red}{-3.67}}$ & 47.79$_{\textcolor{red}{-5.32}}$ & 67.13$_{\textcolor{red}{-2.62}}$ \\
MBCD~\cite{wang2025modality}& $\checkmark$& \ding{55} & \secnd{80.18}$_{\textcolor{blue}{+2.06}}$ & \secnd{76.74}$_{\textcolor{red}{-2.17}}$ & 52.62$_{\textcolor{red}{-0.87}}$ & \secnd{69.85}$_{\textcolor{red}{-0.32}}$ \\
GMP~\cite{li2026balancing} & $\checkmark$& \ding{55} & 78.17$_{\textcolor{blue}{+0.81}}$ & 75.49$_{\textcolor{red}{-0.98}}$ & 51.06$_{\textcolor{red}{-1.27}}$ & 68.24$_{\textcolor{red}{-0.48}}$ \\
\midrule
ERM & \ding{55}& $\checkmark$& 25.96$_{\textcolor{red}{-49.95}}$ & 37.42$_{\textcolor{red}{-40.06}}$ & 22.24$_{\textcolor{red}{-31.16}}$ & 28.54$_{\textcolor{red}{-40.39}}$ \\
RNA-Net~\cite{Planamente_2022_WACV} & \ding{55}& $\checkmark$& 27.90$_{\textcolor{red}{-47.30}}$ & 31.35$_{\textcolor{red}{-46.13}}$ & 18.29$_{\textcolor{red}{-35.29}}$ & 25.85$_{\textcolor{red}{-42.90}}$ \\
SimMMDG~\cite{dong2023SimMMDG}& \ding{55}& $\checkmark$& 30.64$_{\textcolor{red}{-47.95}}$ & 27.59$_{\textcolor{red}{-50.45}}$ & \best{29.23}$_{\textcolor{red}{-26.56}}$ & 29.15$_{\textcolor{red}{-41.66}}$ \\
MOOSA~\cite{dong2024moosa}& \ding{55}& $\checkmark$& 27.04$_{\textcolor{red}{-52.34}}$ & \best{39.07}$_{\textcolor{red}{-39.63}}$ & 22.06$_{\textcolor{red}{-32.72}}$ & 29.39$_{\textcolor{red}{-41.56}}$ \\
CMRF~\cite{fan2024cross}& \ding{55}& $\checkmark$& \best{32.01}$_{\textcolor{red}{-45.93}}$ & \secnd{38.52}$_{\textcolor{red}{-39.74}}$ & 28.03$_{\textcolor{red}{-23.81}}$ & \best{32.85}$_{\textcolor{red}{-36.50}}$ \\
NEL~\cite{zhang2025nonpolarized}& \ding{55}& $\checkmark$& 22.58$_{\textcolor{red}{-53.75}}$ & 28.44$_{\textcolor{red}{-47.98}}$ & 21.01$_{\textcolor{red}{-30.06}}$ & 24.01$_{\textcolor{red}{-43.93}}$ \\
JAT~\cite{li2025towards}& \ding{55}& $\checkmark$& 27.51$_{\textcolor{red}{-50.65}}$ & 33.84$_{\textcolor{red}{-44.15}}$ & 21.65$_{\textcolor{red}{-31.46}}$ & 27.67$_{\textcolor{red}{-42.08}}$ \\
MBCD~\cite{wang2025modality}& \ding{55}& $\checkmark$& \secnd{31.33}$_{\textcolor{red}{-46.79}}$ & 37.96$_{\textcolor{red}{-40.95}}$ & \secnd{28.69}$_{\textcolor{red}{-24.80}}$ & \secnd{32.66}$_{\textcolor{red}{-37.51}}$ \\
GMP~\cite{li2026balancing} & \ding{55}& $\checkmark$& 29.41$_{\textcolor{red}{-47.95}}$ & 34.20$_{\textcolor{red}{-42.27}}$ & 23.86$_{\textcolor{red}{-28.47}}$ & 29.16$_{\textcolor{red}{-39.56}}$ \\
\bottomrule
\end{tabular}
\end{threeparttable}
}
\label{tab:hac-dg-missing}
\end{table*}

\section{Compute Resources}
All experiments were conducted on servers equipped with NVIDIA RTX 3090 and RTX 4090 GPUs. Each model was trained using standard deep learning frameworks with GPU acceleration. In total, 7,402 neural networks were trained across 95 cross-domain tasks, reflecting the large computational scale of MMDG-Bench.



\end{document}